\documentclass[sigconf,author]{acmart}

\renewcommand\footnotetextcopyrightpermission[1]{} 
\usepackage{booktabs} 
\setcopyright{rightsretained}
\usepackage{graphicx}
\usepackage{color}
\usepackage{soul}
\usepackage{multirow}

\graphicspath{{./figures/}}

\usepackage{algorithm}
\usepackage{algpseudocode}
\algnewcommand\algorithmicforeach{\textbf{for each}}
\algdef{S}[FOR]{ForEach}[1]{\algorithmicforeach\ #1\ \algorithmicdo}
\usepackage{algpseudocode}
\algrenewcommand\textproc{}

\let\oldReturn\Return
\renewcommand{\Return}{\State\oldReturn}

\usepackage[caption=false]{subfig}

\makeatletter
\newenvironment{subfigures}
 {\begin{minipage}{\columnwidth}\def\@captype{figure}\centering}
 {\end{minipage}}
\makeatother

\begin{document}

\copyrightyear{2019} 
\acmYear{2019} 
\setcopyright{acmcopyright}
\acmConference[GECCO '19]{Genetic and Evolutionary Computation Conference}{July 13--17, 2019}{Prague, Czech Republic}
\acmBooktitle{Genetic and Evolutionary Computation Conference (GECCO '19), July 13--17, 2019, Prague, Czech Republic}
\acmPrice{15.00}
\acmDOI{10.1145/3321707.3321723}
\acmISBN{978-1-4503-6111-8/19/07}

\title{Learning with Delayed Synaptic Plasticity}


\author{Anil Yaman}
\affiliation{%
  \institution{Eindhoven University of Technology}
  \city{Eindhoven} 
  \state{the Netherlands} 
}
\email{a.yaman@tue.nl}

\author{Giovanni Iacca}
\affiliation{%
  \institution{University of Trento}
  \city{Trento} 
  \state{Italy} 
}
\email{giovanni.iacca@unitn.it}

\author{Decebal Constantin Mocanu}
\affiliation{%
  \institution{Eindhoven University of Technology}
  \city{Eindhoven} 
  \state{the Netherlands} 
}
\email{d.c.mocanu@tue.nl}

\author{George Fletcher}
\affiliation{%
  \institution{Eindhoven University of Technology}
  \city{Eindhoven} 
  \state{the Netherlands} 
}
\email{g.h.l.fletcher@tue.nl}

\author{Mykola Pechenizkiy}
\affiliation{%
  \institution{Eindhoven University of Technology}
  \city{Eindhoven} 
  \state{the Netherlands} 
}
\email{m.pechenizkiy@tue.nl}

\renewcommand{\shortauthors}{A. Yaman et al.}

\begin{abstract}

The plasticity property of biological neural networks allows them to perform learning and optimize their behavior by changing their configuration. Inspired by biology, plasticity can be modeled in artificial neural networks by using Hebbian learning rules, i.e. rules that update synapses based on the neuron activations and reinforcement signals. However, the distal reward problem arises when the reinforcement signals are not available immediately after each network output to associate the neuron activations that contributed to receiving the reinforcement signal. In this work, we extend Hebbian plasticity rules to allow learning in distal reward cases. We propose the use of neuron activation traces (NATs) to provide additional data storage in each synapse to keep track of the activation of the neurons. Delayed reinforcement signals are provided after each episode relative to the networks' performance during the previous episode. We employ genetic algorithms to evolve delayed synaptic plasticity (DSP) rules and perform synaptic updates based on NATs and delayed reinforcement signals. We compare DSP with an analogous hill climbing algorithm that does not incorporate domain knowledge introduced with the NATs, and show that the synaptic updates performed by the DSP rules demonstrate more effective training performance relative to the HC algorithm.

\end{abstract}



\maketitle

\section{Introduction}

The plasticity property of biological neural networks enables the networks to change their configuration (i.e. \textit{topology} and/or \textit{connection weights}) and learn to perform certain tasks during their lifetime. The learning process involves searching through the possible configuration space of the networks until a configuration that achieves a satisfactory performance is reached. Modelling plasticity, or rather \emph{evolving} it, has been a long-standing goal in Neuro-evolution (NE), a research field that aims to design artificial neural networks (ANNs) using evolutionary computing approaches~\cite{stanley2019,floreano2008neuroevolution,kow2016growing}.

Neuro-evolutionary methods can be divided roughly in \textit{direct} and \textit{indirect} encoding approaches~\cite{floreano2008neuroevolution}. 
In direct encoding, the topology and/or connection weights of the ANNs are directly represented within the genotype of the individuals. However, the number of possible network configurations (i.e. connectivity) increases exponentially depending on the number of neurons in a network. Therefore, it may be challenging to scale direct encoding approaches to large networks~\cite{yaman2018limited,mocanu2018scalable}. In indirect encoding on the other hand, this drawback is overcome by encoding in the genotype rather than the network parameters \emph{per se}, the rules to construct and/or optimize the ANNs, usually during their lifetime~\cite{kow2016growing,soltoggio2017born}. Indirect encoding has also the additional advantage of being biologically plausible, as evidence has showed that biological neural networks undergo changes throughout their lifetime without the need to change the genes involved in the expression of the neural networks.

Among the indirect encoding approaches, evolving plastic artificial neural networks (EPANNs)~\cite{coleman2012evolving,soltoggio2017born} implement plasticity by modifying the networks' configuration based on some plasticity rules which are activated throughout the networks' lifetime. These are encoded within the genotype of a population of individuals, in order to optimize the learning procedure by means of evolutionary computing approaches. One possible way of modelling plasticity rules in EPANNs is by means of \textit{Hebbian learning}, a biologically plausible mechanism hypothesized by Hebb~\cite{hebb1949} to model synaptic plasticity~\cite{kuriscak2015}. According to this model, synaptic updates are performed based on local neuron activations. Previous works on EPANNs have implemented this concept by using evolutionary computing to optimize the coefficients of first-principle equations that describe the synaptic updates between pre- and post-synaptic neurons~\cite{coleman2012evolving,floreano2000evolutionary,niv2002evolution}. Others have used machine learning models (i.e. ANNs) that can take the activations of pre- and post-synaptic neurons as inputs to compute the synaptic update as output~\cite{risi2010indirectly,orchard2016evolution}. 

Reinforcement signals (and/or the output of other neurons, as in neuromodulation schemes~\cite{soltoggio2008evolutionary,durr2008evolvability}) can be used as modulatory signals to guide the learning process by signaling how and when the plasticity rules are used. To allow learning, these signals are usually required right after each output of the network, to help associating the activation patterns of neurons with the output. If the reinforcement signals are available only after a certain period of time, but not immediately after each action step, it may not be possible to directly associate the behavior of the network to the activation patterns of the neurons. Thus, the \textit{distal reward problem}~\cite{soltoggio2013solving,izhikevich2007solving} arises, which cannot be addressed using Hebbian learning in its basic form, but requires a modified learning model to take into account the neuron activations over a certain period of time.

In this work, we propose a modified Hebbian learning mechanism, which we refer to as \textit{delayed synaptic plasticity (DSP)}, for enabling plasticity in ANNs in tasks with distal rewards. We introduce the use of \textit{neuron activation traces (NATs)}, i.e. additional data storage in each synapse that keep track of the average pre- and post-synaptic neuron activations. We use discrete DSP rules to perform synaptic updates based on the NATs and a reinforcement signal provided after a certain number of activation steps, that we refer to as an \emph{episode}. The reinforcement signals are based on the performance of the agent relative to the previous episode (i.e. if the agent performs better/worse relative to the previous episode, a positive/negative reinforcement signal is provided). We further introduce competition in incoming synapses of neurons to stabilize delayed synaptic updates and encourage self-organization in the connectivity. As such, the proposed DSP scheme is a distributed and self-organized form of learning which does not require global information of the problem.

We employ \textit{genetic algorithms (GA)} to evolve DSP rules to perform synaptic changes on \textit{recurrent neural networks (RNNs)} that have to learn to navigate in a triple T-maze to reach a goal location. To test the robustness of the evolved DSP rules, we evaluate them for multiple trials with various goal positions. We then note how the process of training RNNs for a task using DSP can be seen as analogous to optimizing them using a single-solution metaheuristic, except for the fact that in contrast to general-purpose metaheuristics, the DSP rules take into account the (domain-specific) knowledge of the local neuron interactions for updating the synaptic weights. Therefore, to assess the effect of the domain knowledge introduced with DSP, we compare our results with a classic \textit{hill climbing (HC)} algorithm. Our results show that DSP is highly effective in speeding up the optimization of RNNs relative to HC in terms of number of function evaluations needed to converge. On the other hand, the NATs data structure introduces an additional computational complexity.

The rest of the paper is organized as follows: in Section~\ref{sec:background}, we discuss Hebbian learning and the distal reward problem; in Section~\ref{sec:methods}, we introduce our proposed approach for DSP and provide a detailed description of the evolutionary approach we used to optimize DSP; in Section~\ref{sec:experimentalSetup}, we present our experimental setup; in Section~\ref{sec:results}, we provide a comparison analysis of our proposed approach and the baseline HC algorithm; finally, in Section~\ref{sec:conclusion}, we discuss our conclusions.

\vspace{-0.1cm}

\section{Background} \label{sec:background}

ANNs are computational models inspired by biological neural networks~\cite{de2006fundamentals}. They consists of a number of \textit{artificial neurons} arranged in a certain connectivity pattern. Adopting the terminology from biology, a directional connection between two neurons is referred to as a \textit{synapse}, and the antecedent and subsequent neurons relative to a synapse are called \textit{pre-synaptic} and \textit{post-synaptic neurons}. The activation of a post-synaptic neuron $a_i$ can be computed by using the following formula:
\begin{equation}
a_i = \psi \big( \sum_j w_{i,j} \cdot a_j \big)
\label{eq:annActivation}
\end{equation}
where $a_j$ is the activation of the $j$-th pre-synaptic neuron, $w_{i,j}$ is the synaptic efficiency between $i$-th and $j$-th neurons, and $\psi(\cdot)$ is an activation function (i.e. a sigmoid function~\cite{de2006fundamentals}). The pre-synaptic neuron $a_0$ is usually set a constant value of 1.0 and referred to as the \textit{bias}. Let us introduce also the equivalent matrix notation, that will be used in the following sections: considering that all the activations of pre- and post-synaptic neurons in layers $k$ and $l$ can be represented, respectively, as a column vectors $\boldsymbol{A}_k$ and $\boldsymbol{A}_l$, then the activation of post-synaptic neurons can be calculated as: $\boldsymbol{A}_l = \psi \big( \boldsymbol{W}_{lk} \cdot \boldsymbol{A}_k\big)$, where $\boldsymbol{W}_{lk}$ is a matrix that encodes all the synaptic weights.

\vspace{-0.1cm}
\subsection{Hebbian Learning}

Hebbian learning is a biologically plausible learning model that performs synaptic updates based on local neuron activations~\cite{kuriscak2015}. In its general form, a synaptic weight $w_{i,j}$ at time $t$ is updated using the \textit{plain} Hebbian rule, i.e.: 
\begin{equation}
     w_{i,j}(t+1) = w_{i,j}(t) + m(t) \cdot \Delta w_{i,j}(t)
\end{equation}
\begin{equation}
    \Delta w_{i,j}(t) = f \big( a_i(t),a_j(t) \big) = \eta \cdot a_i(t) \cdot a_j(t)
\end{equation}
where $\eta$ is the learning rate and $m(t)$ is the modulatory signal. 

Using the plain Hebbian rule, the synaptic efficiency between neurons $i$ and $j$, $w_{i,j}$ is strengthened/weakened when the sign of their activations are positively/negatively correlated, and does not change when there is no correlation. The modulatory signal $m(t)$ can alter the sign of Hebbian learning implementing \textit{anti-Hebbian} learning if the modulatory signal is negative~\cite{brown1990hebbian,soltoggio2012modulated}.

The plain Hebbian rule may cause indefinite increase/decrease in the synaptic weights because strengthened/weakened synaptic weights increase/decrease neuron correlations, which in turn cause synaptic weights to be further strengthened/weakened. Several versions of the rule were suggested to improve its stability~\cite{vasilkoski2011review}.

\vspace{-0.1cm}
\subsection{Evolving Plastic Artificial Neural Networks}

Plasticity in EPANNs makes them capable of learning during their lifetime by changing their configuration~\cite{soltoggio2017born,coleman2012evolving,mouret2014artificial}. Hebbian learning rules are often employed to model plasticity in EPANNs. On the other hand, the evolutionary aspect of EPANNs typically involve the use of evolutionary computing approaches to find a near-optimum Hebbian learning rules to perform synaptic updates.

In the literature, several authors suggested evolving the type and parameters of Hebbian learning rules using evolutionary computing. For instance, Floreano and Urzelai evolved four Hebbian learning rules and their parameters in an unsupervised setting where synaptic changes are performed periodically during the networks' lifetime~\cite{floreano2000evolutionary}. Niv \textit{et al.,}~\cite{niv2002evolution} evolved the parameters of a Hebbian rule that defines a complex relation between the pre- and post-synaptic neuron activations on a reinforcement learning setting. Others suggested evolving complex machine learning models (i.e. ANNs) to replace Hebbian rules to perform synaptic changes~\cite{runarsson2000evolution,izquierdo2007hebbian}. Orchard and Wang~\cite{orchard2016evolution} compared evolved plasticity based on a parameterized Hebbian learning rule with evolved synaptic updates based on ANNs. However, they also included the initial weights of the synapses into the genotype of the individuals. Risi and Stanley~\cite{risi2010indirectly} used \textit{compositional pattern producing networks} (CPPN)~\cite{stanley2007compositional} to perform synaptic updates based on the location of the connections. Tonelli and Mouret~\cite{tonelli2013relationships} investigated the learning capabilities of plastic ANNs evolved using different encoding schemes.

\vspace{-0.1cm}

\subsection{Distal Reward Problem}

When reinforcement signals are available after a certain period of time, it may not be possible to associate the neuron activations that contributed to receiving the reinforcement signals. This is referred to as the \emph{distal reward problem}, and has been studied in the context of time dependent plasticity~\cite{gerstner2018eligibility,izhikevich2007solving,soltoggio2013solving}.

From a biological viewpoint, \textit{synaptic eligibility traces (SETs)} have been suggested as a plausible mechanism to trace the activations of neurons over a certain period of time~\cite{gerstner2018eligibility} by means of chemicals present in the synapses. According to this mechanism, co-activations of neurons may trigger an increase in the SETs, which then decay over time. Therefore, their level can indicate a recent co-activation of neurons when a reinforcement signal is received, and as such be used for distal rewards.

\vspace{-0.1cm}

\section{Methods}\label{sec:methods}

In this work, we focus on a navigation in triple T-maze environment (see Section~\ref{sec:experimentalSetup}) that requires memory capabilities. Thus, we use a RNN model illustrated in Figure \ref{fig:rnnArchitecture}.

Our RNN model consist of input, hidden and output layers connected with four sets of connections. All the neuron activations are set to zero at the initial time step $t = 0$. Then the activation of the neurons in the hidden and output layers at each discrete time step $t+1$, respectively $\boldsymbol{A}_{h}(t+1)$ and $\boldsymbol{A}_{o}(t+1)$, are updated as:
\begin{equation}
\begin{split}
\boldsymbol{A}_h(t+1) = & \psi 
\Big( \boldsymbol{W}_{hi} \cdot \boldsymbol{A}_i(t+1) + \alpha_h \cdot \boldsymbol{W}_{h} \cdot \boldsymbol{A}_h(t) \\ & + \alpha_o \cdot \boldsymbol{W}_{ho} \cdot \boldsymbol{A}_o(t) \Big)
\end{split}
\label{eq:hiddenActivation}
\end{equation}
\begin{equation}
\boldsymbol{A}_o(t+1) = \psi 
\Big( 
\boldsymbol{W}_{oh} \cdot \boldsymbol{A}_h(t+1)
\Big)
\label{eq:outputActivation}
\end{equation}
where:\\
1) $\boldsymbol{W}_{hi}$ and $\boldsymbol{W}_{oh}$ are feed-forward connections between input-hidden and hidden-output layers, $\boldsymbol{W}_{h}$ is the recurrent connection of the hidden layer, and $\boldsymbol{W}_{ho}$ is the feedback connection from the output layer to hidden layer. The recurrent and feedback connections provide inputs of the activations of the hidden and output neurons from the previous time step. We do not allow self-recurrent connections in $\boldsymbol{W}_{h}$ (the diagonal elements of $\boldsymbol{W}_{h}$ equal to zero).\\
2) $\alpha_h$ and $\alpha_o$ are coefficient used to scale the recurrent and feedback inputs from the hidden and output layers respectively.\\
3) $\psi(\cdot)$ is a binary activation function defined as:
\begin{equation}
\psi(x) = \left\{
 \begin{array}{lc}
  1, & \mbox{if }x>0\mbox{;}\\
  0, & \mbox{otherwise.}
 \end{array}
\right. 
\label{eq:activationFunction}
\end{equation}
\subsection{Learning} \label{sec:learning}
The learning process of an RNN is a search process that aims to find an optimal configuration of the network (i.e., its synaptic weights) that can map the sensory inputs to proper actions in order to achieve a given task. In this work, we use independently the proposed DSP rules, and the HC algorithm~\cite{de2006fundamentals}, to perform the learning the RNNs for the specific tripe T-maze navigation task, and compare their results. Both of these approaches perform synaptic updates based on the progress of the performance of an individual agent in consecutive episodes; however, the DSP rules incorporate the knowledge of the local neuron interactions, while HC uses a certain random perturbation heuristic that does not incorporate any knowledge. Here, we assume that the progress of the performance of an agent can be measured relative to its performance in a previous episode. We refer to the measure of the performance of an agent in a single episode as ``episodic performance'' (EP). We should note that we formalize our task as a minimization problem. Therefore, in our experiment an agent with a lower EP is better.
\begin{figure}[t]
\begin{center}
\begin{subfigures}
\subfloat[]{\includegraphics[width=\columnwidth]{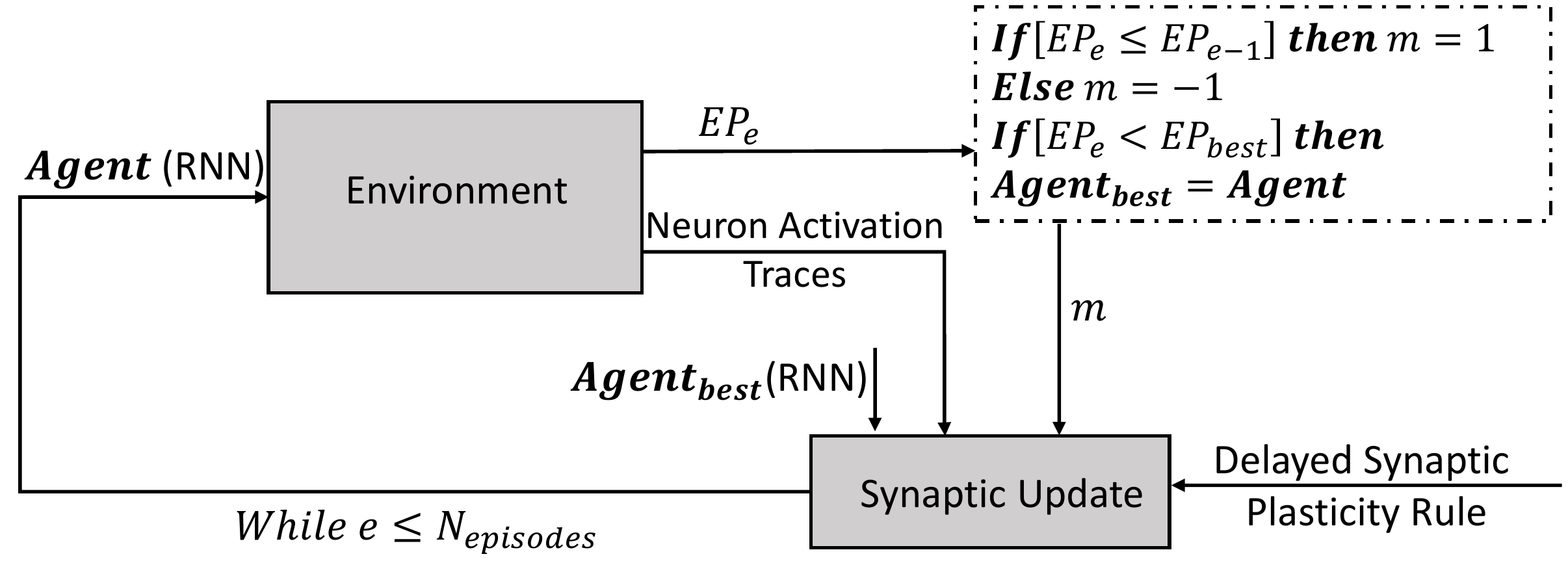}\label{fig:framework}}
\end{subfigures}
\subfloat[]{\includegraphics[width=\columnwidth]{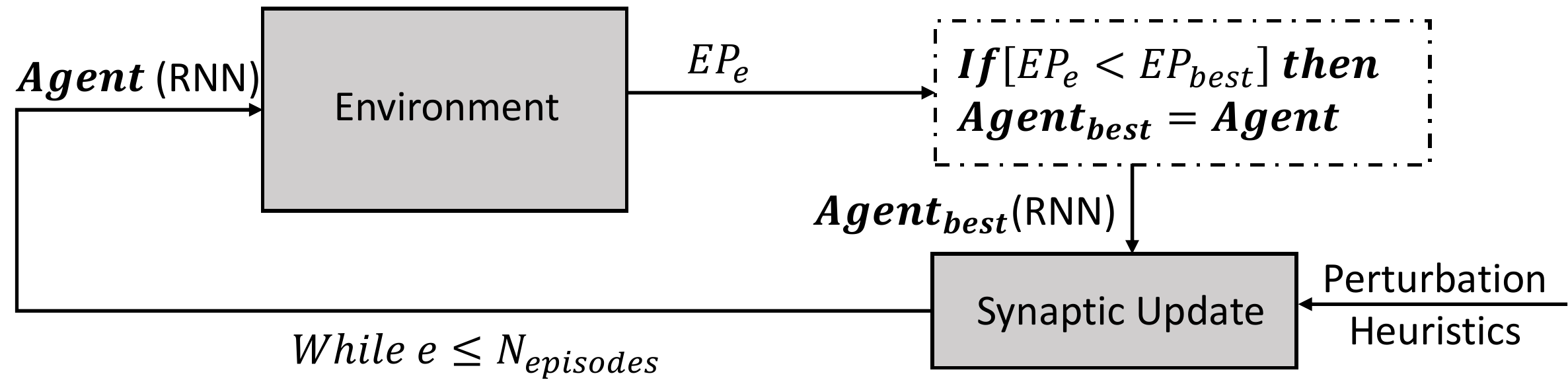}\label{fig:hillClimbingFramework}}
\end{center}
\caption{(a) The learning process by using the delayed synaptic plasticity, and (b) the learning process by optimizing the parameters of the RNNs using the hill climbing algorithm.} \label{fig:learningArchitecture}
\end{figure}

The illustration of the optimization processes of the RNNs using the DSP rules and the HC algorithm are given in Figures~\ref{fig:framework} and \ref{fig:hillClimbingFramework} respectively. Both algorithms run for a pre-defined number of episodes ($N_{episodes}$), starting from an RNN with randomly initialized synaptic weights, and record the best encountered RNN throughout the optimization process. 

In the DSP-based algorithm illustrated in Figure~\ref{fig:framework}, the synaptic updates are performed after each episode ---thus, the synaptic weights of the RNN during an episode are fixed--- using DSP rules which take the RNN, NATs, and a modulatory signal as inputs. The NATs provide the average interactions of post- and pre-synaptic neurons during an episode. The structure of the NATs alongside with the DSP rules are explained in Section~\ref{sec:DSP} in detail. The modulatory signal is used as a reinforcement which depends on the performance of the agent in the current episode relative to its performance during the previous episode. If the current episodic performance $EP_e$ is lower than the previous episodic performance $EP_{e-1}$, then the modulatory signal $m$ is set to $1$ (reward), otherwise $m$ is set to $-1$ (punishment).    

The HC algorithm that is illustrated in Figure~\ref{fig:hillClimbingFramework} performs instead synaptic updates after each episode using a perturbation heuristic that does not assume any knowledge of the neuron activations. We use a Gaussian perturbation with a zero mean and unitary standard deviation and scale it with a parameter $\sigma$ to perturb all the synaptic weights of the RNN of the best agent. The synaptic update procedure generates a new candidate $\boldsymbol{Agent}$ that is tested in the environment for the next episode and replaced with the best RNN if it performs better. Conventionally, in standard HC the measure of the performance of an agent in an episode would be called ``fitness''. However, in this work we refer to it as EP, to make it analogous to the algorithm that uses DSP rules.

\vspace{-0.12cm}
\subsection{Evolving Delayed Synaptic Plasticity} \label{sec:DSP}
We propose delayed synaptic plasticity to allow synaptic changes based on the progress of the performance of an agent relative to its performance during the previous episode, i.e.:
\begin{equation}
\Delta w_{i,j}(e) = DSP(NAT_{i,j},m, \theta)
\label{eq:dspSynapticChange}
\end{equation}
\begin{equation}
w_{i,j}^{\prime}(e) = w_{i,j}(e) + \eta \cdot \Delta w_{i,j}(e)
\label{eq:episodicSynapticChange}
\end{equation}
where the synaptic change $\Delta w_{i,j}(e)$ between neurons $i$ and $j$ after an episode $e$ is computed based on their NATs, the modulatory signal $m$ (which can either be $1$ or $-1$, see Section~\ref{sec:learning}), and a threshold $\theta$ that is used to convert NATs into binary vectors. The resulting DSP synaptic changes can be of three types (\textit{decreased}, \textit{stable}, or \textit{increased}), i.e. at each time step $DSP(NAT_{i,j},m, \theta) \in \{-1,0,1\}$. In Equation \eqref{eq:episodicSynapticChange}, the synaptic change is scaled with a learning rate $\eta$.

Subsequent to the synaptic updates of all synapses, the synaptic weights for the next episode $w_{i,j}(e+1)$ are computed as:
\begin{equation}
w_{i,j}(e+1) = \frac{w^{\prime}_{i,j}(e)}{|| \boldsymbol{w}^{\prime}_{i}(e) ||_2}  \label{eq:normEquation}
\end{equation}
where $ \boldsymbol{w}^{\prime}_{i}(e)$ is a row vector encoding all incoming synaptic weights of a post-synaptic neuron $i$ and $|| \cdot ||_2$ is the Euclidean norm. Here, the synaptic weights are scaled to have a unitary length, thus preventing an indefinite increase/decrease, and allowing self-organized competition between synapses~\cite{el2018locally}. Alternatively, a decay mechanism and/or saturation limits for weights can be introduced to prevent an indefinite increase/decrease of the weights~\cite{soltoggio2012modulated}.
\begin{figure}[ht!]
\begin{center}
\includegraphics[width = 0.75\columnwidth]{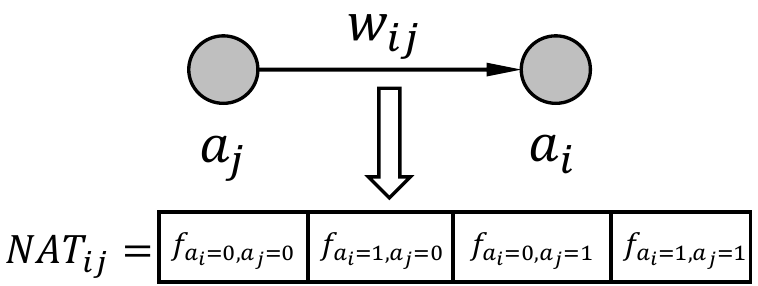}
\caption{The neuron activation trace $NAT_{i,j}$ of the pre- and post-synaptic neuron activations $a_j$ and $a_i$.}
\label{fig:NAT}
\end{center}
\end{figure}

The NAT data structure is illustrated in Figure~\ref{fig:NAT}. It keeps track of the frequencies of the pre- and post-synaptic neuron activation states throughout an episode. We use four-dimensional vectors for each synapse ---since there are four possible states for the activations of pre- and post-synaptic neurons--- to store the number of times the pre- and post-synaptic neurons were in one of the following states: $00$, $01$, $10$, and $11$, where the first and second bits represent the pre- and post-synaptic neuron activations, and $0$ and $1$ represent non-active and active states of neurons respectively. At the beginning of an episode, all NATs are initialized as zeros; and the end of an episode, they are divided by the number of total activations to convert them into frequencies. 

The NATs are used in their binary forms to discretize the search space. Each frequency in a NAT is converted to either $0$ or $1$ based on a threshold $\theta$ ($1$ for the frequencies more than $\theta$, and $0$ for the ones that are below $\theta$). Thus, a DSP rule, as illustrated in Table~\ref{tab:DSPrule}, is a combined form of 32 synaptic change decisions provided for all possible states of the binary forms of the NATs and the signal $m$.

\begin{table}[t]
\caption{A delayed synaptic plasticity rule visualized in a tabular form. Depending on a certain threshold $\theta$ for converting the binary forms of specific NATs, the DSP rule provides the synaptic changes $x_1, x_2, \hdots, x_{32}$ for all possible combinations of binary NATs and $m$ states.}\label{tab:DSPrule}
\begin{tabular}{|c|c|c|c|c|c|}
\hline
\multicolumn{4}{|c|}{$\boldsymbol{NAT^{\theta}}$}                                                                 & \multirow{2}{*}{$\boldsymbol{m}$} & \multirow{2}{*}{$\boldsymbol{\Delta w}$} \\ \cline{1-4}
$\boldsymbol{00}$             & $\boldsymbol{01}$            & $\boldsymbol{10}$           & $\boldsymbol{11}$            &                             &                   \\ \hline\hline
$0$                       & $0$                      & $0$                      & $0$                     & $-1$                          & $x_1$                 \\ \hline
$0$                       & $0$                     & $0$                      & $0$                      & $1$                          &   $x_2$                 \\ \hline
$\hdots$                     & $\hdots$                    & $\hdots$                     & $\hdots$                     & $\hdots$                    & $\hdots$                \\ \hline
\multicolumn{1}{|c|}{1} & \multicolumn{1}{c|}{1} & \multicolumn{1}{c|}{1} & \multicolumn{1}{c|}{1} & \multicolumn{1}{c|}{1}      &             $x_{32}$       \\ \hline
\end{tabular}
\end{table}

We use genetic algorithms (GAs) to evolve three possible synaptic changes $\Delta w = \{-1, 0 ,1\}$ (\textit{decrease, stable, increase}) for each possible states of the NATs and $m$. In total, there are 32 possible states that can take three possible values. Thus, there is a total of $3^{32}$ number of possible DSP rules. In addition to the discrete part, we optimize four continuous parameters ($\eta, \theta, \alpha_h, \alpha_o$) by including them into the genotype of the individuals. We used suitable evolutionary operators separately for discrete and continuous variables. The details of the GA are provided in Section~\ref{sec:GA}.

\vspace{-0.15cm}
\section{Experimental Setup} \label{sec:experimentalSetup}

In the following sections, we provide the details of our experimental setup including the test environment, the architecture of the agent, and the details of the GA.

\vspace{-0.15cm}
\subsection{Triple T-Maze Environment}
\begin{figure}[ht!]
\begin{center}
\includegraphics[width = 0.8\columnwidth]{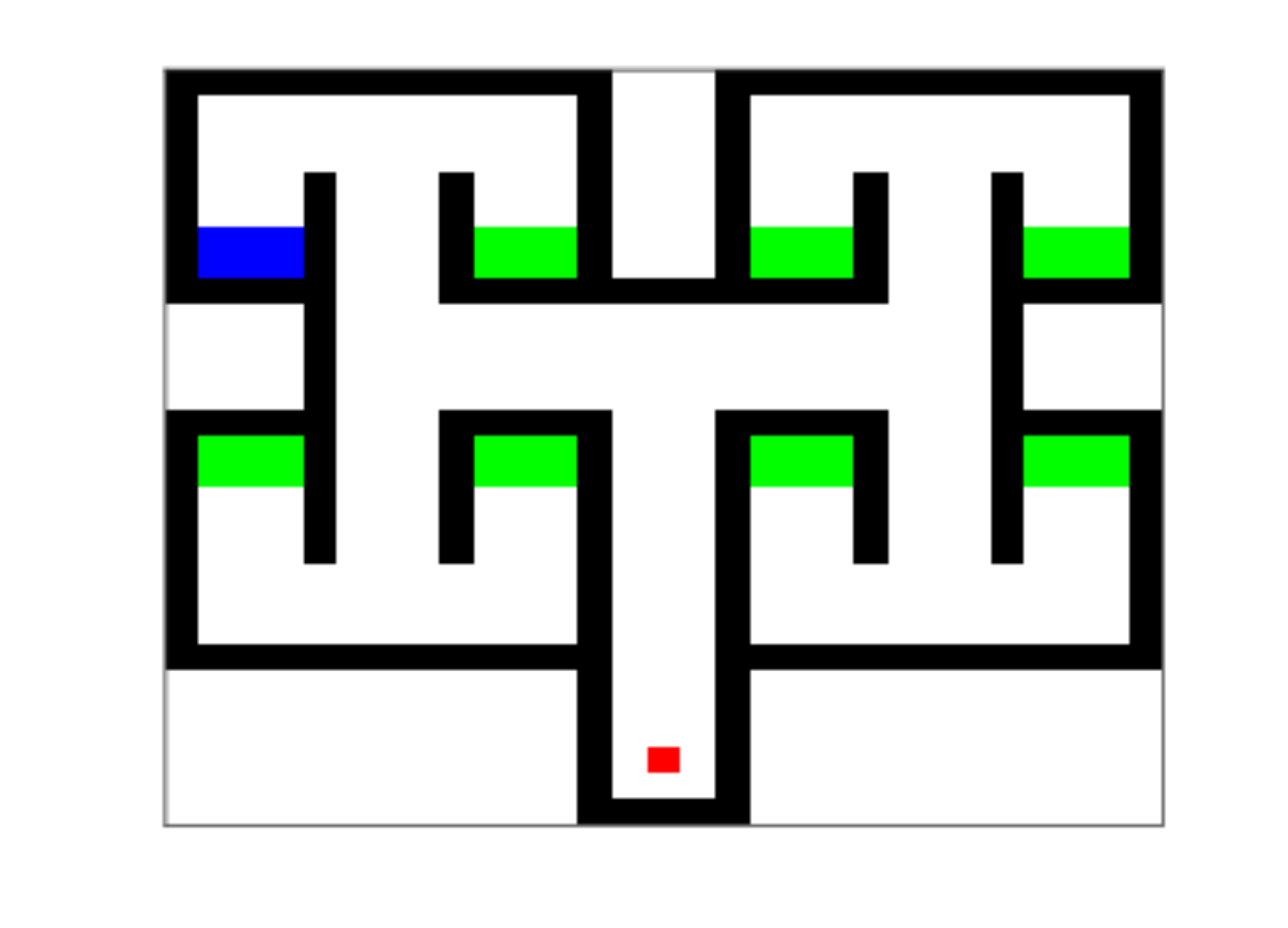}
\caption{Triple T-maze environment. The walls, starting position of the agent, goal and pits are color-coded using black, red, blue and green.}
\label{fig:tripleTMaze}
\end{center}
\end{figure}
A visual illustration of the triple T-maze environment is given in Figure~\ref{fig:tripleTMaze}. The triple T-maze environment consists of $29 \times 29$ cells. Each cell can be occupied by one of five possibilities: \textit{empty, wall, goal, pit, agent}, color-coded in white, black, blue, green, red respectively. The starting position of the agent is illustrated in red. There are eight final positions, illustrated in blue or green. Among eight final positions, one of them is assigned as the goal position (in blue), and the rest as pits (in green). Starting from the initial position, an agent navigates the environment for a total of 100 action steps. To solve the task, it is required to find a set of connection weights of the recurrent neural network that can allow the agent to achieve the goal from the starting position.

\subsection{Agent Architecture}

\begin{figure}[t]
\begin{center}
\begin{subfigures}
\subfloat[]{\includegraphics[width=0.75\columnwidth]{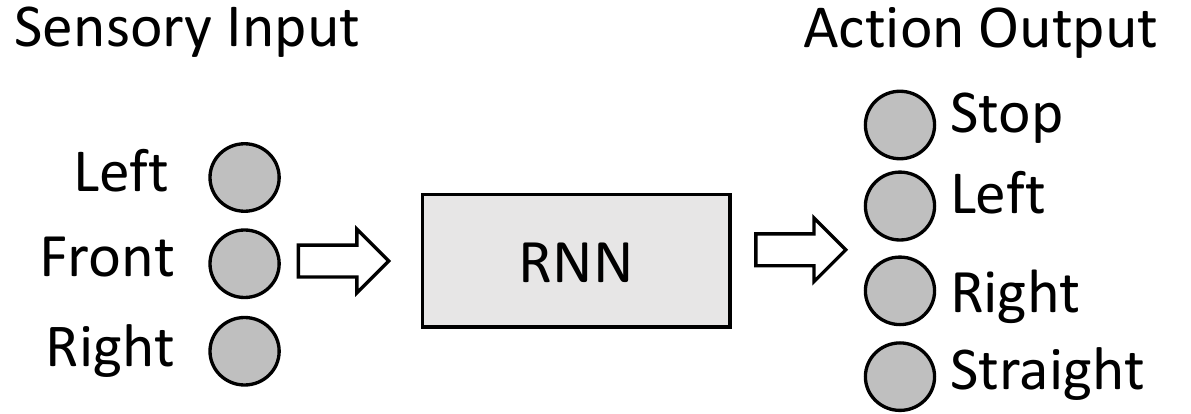}\label{fig:agent}}
\\
\subfloat[]{\includegraphics[width=0.75\columnwidth]{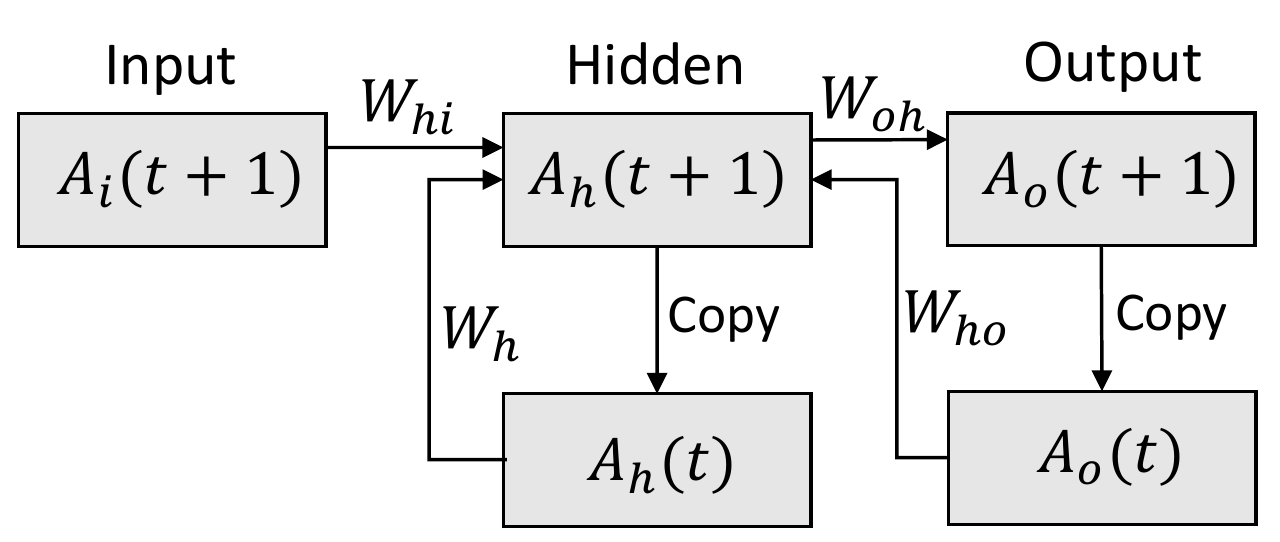}\label{fig:rnnArchitecture}}
\end{subfigures}
\end{center}
\caption{ (a) The sensory inputs and action outputs of the recurrent neural networks and (b) the architecture of the networks that are used to control the agents.} \label{fig:agentArchitecture}
\end{figure}

An illustration of the architecture of the agents used for the triple T-maze environment is provided in Figure~\ref{fig:agentArchitecture}. An agent has a certain orientation in the environment, facing one of the four possible directions along the $x$ and $y$ axes of the environment (i.e. north, south, west, east), and can only move one cell at a time horizontally or vertically. It can take sensory inputs from the nearest cells from its left, front and right, depending on its orientation. We let the agent sense only an empty cell, or a wall (i.e., the agent is not aware of the presence of the goal or pits), so the sensory inputs are encoded using one bit, representing an empty cell with $0$ and a wall with $1$. Thus, there are three bits as input, that the agent uses to decide one of four possible actions: \textit{stop}, \textit{left}, \textit{right}, and \textit{straight}. In case of stop, the agent stays in the same cell with the same orientation for an action step; in cases of left and right, the agent changes its orientation accordingly and proceeds one cell straight; in case of straight, the agent keeps its original orientation and proceeds one cell straight. If the cell the agent wants to move in is occupied with wall, then the agent stays in its original position.
We use RNNs to control the agents since our task requires memory capabilities, and may not be solved by a static architecture such as a feed-forward ANN. The recurrent and feedback connections between the hidden-hidden and output-hidden layers help agents to keep track of the sequences of actions they perform throughout an episode. The RNN networks used in all experiments consist of 20 hidden neurons unless otherwise is specified. Therefore, the networks consist of $(3+1) \times 20 = 80$ input to hidden connections, $20\times 19 = 380$ hidden to hidden connections (except self), $(20+1)\times 4 = 84$ hidden to output connections, and $4 \times 20 = 80$ output to hidden connections, in total of $624$ connections between all layers ($+1$ refers to the bias).

\subsection{Genetic Algorithm}\label{sec:GA}

We use a standard GA to evolve DSP rules with the exception of a custom-designed mutation operator. The genotype of the DSP rules consists of 32 discrete ($\Delta w$ for all possible states of the NATs and $m$, see Section~\ref{sec:DSP}) and four continuous values ($\eta\in [0,1],\theta \in [0,1],\alpha_h\in [0,1],\alpha_o\in [0,1]$). Therefore, we encode a population of individual genotypes represented by 36-dimensional discrete/real-valued vectors. 

The pseudo-code for the evaluation function is provided in Algorithm~\ref{alg:evaluation}. To find the DSP rules that can robustly learn the triple T-maze navigation task independently of the goal position and starting from a random RNN state, each evaluation called during the evolutionary process consists of testing each rule for five trials for each of the eight possible goal positions (for total of 40 independent trials). The final positions (one goal in blue, seven pits in green) of the triple T-maze environment are shown in Figure~\ref{fig:tripleTMaze}. We switch the position of the goal with one of the pits to have eight distinct goal positions in total. For all distinct goal positions, the rest of the final positions are assigned as pits. Due to the computational complexity, the maximum number of episodes per trial is set to 100. 

The $EP$ of an agent in an episode is computed as follows:
\begin{equation}
EP = \left\{
 \begin{array}{lr}
  steps(Agent), & \mbox{if the goal is reached;}\\
  N_{steps} + d(XY(Agent),XY(g)), & \mbox{Otherwise.}
 \end{array}
\right. 
\label{eq:EP}
\end{equation}
For each episode, the agent is given 100 action steps ($N_{steps}$) to reach the goal. We aim to minimize the number of action steps that the agents take to reach the goal (in case they do reach it); otherwise, we aim to minimize their distance to the goal. Thus, if the agent reaches the goal, the EP is the number of action steps the agent took to reach to the goal ($steps(Agent)$). Otherwise, the $EP$ is $N_{steps} + d(XY(Agent),XY(g))$, that is the maximum number of total action steps, plus the distance between the final agent's position and the goal's position. The distance between the agent and the goal is found by the A* algorithm~\cite{zeng2009finding}, which finds the closest path (distance) taking into account the obstacles. Additionally, the EP of an agent is increased by 5 every time the agent steps into a pit (recalling that the $EP$ is minimized).

\begin{algorithm}[t]
	\begin{algorithmic}[1]
	    \Procedure{Evaluate}{$\boldsymbol{x}, trials, N_{episodes}$}
	        \State{$fitness \gets 0$}
            \ForEach {$t \in trials$} 
                \ForEach {$g \in goalPositions$}\Comment{$N_{goals}$ positions} 
                    \State $EP_0\gets inf$, $EP_{best}\gets inf$, $e\gets 1$
                    \State $RNN\gets initializeRandom$
                    \While {$e \leq N_{episodes}$} 
                        \State $EP_e\gets testNetwork(RNN)$
                        \State{$m\gets-1$}
                        \If{$EP_e \leq EP_{e-1}$}
                            \State{$m\gets 1$}
                        \EndIf
                        \State{$EP_{best}\gets min(EP_e, EP_{best})$}
                        \State{$RNN \gets synapticUpdate(RNN,\boldsymbol{x},m)$}
                        \State{$e\gets e+1$}
               	    \EndWhile
               	    \State{$fitness \gets fitness + EP_{best}$}
               	\EndFor
            \EndFor
            \Return $fitness \big/ (trials \cdot N_{goals})$
		\EndProcedure
	\end{algorithmic}
\caption{Evaluation of an individual DSP rule}\label{alg:evaluation}
\end{algorithm}

Finally, the fitness value (the lower the better) of a DSP rule is obtained by averaging $EP_{best}$ found in each trial (i.e. for 5 trials and 8 distinct goal positions, in total of 40 trials). We should note that the proposed design of the $EP$ and fitness is chosen to introduce a gradient into the evaluation process of agents. For instance, the selection process is likely to prefer agents that on average reach the goal with a smaller number of action steps, and with the least number of interactions with pits.

To limit the runtime of the algorithm, we use a small population size, 14 individuals in total. We employ a \textit{roulette wheel selection} operator with an elite number of four, which copies the four individuals with the highest fitness values to the next generation without any change. The rest of the individuals are selected with a probability proportional to their fitness values. We use \textit{1-point crossover} operator with a probability of 0.5. We designed a custom \textit{mutation} operator which re-samples each discrete dimension of the genotype with a probability of 0.15, and performs a Gaussian perturbation with zero mean and 0.1 standard deviation for the continuous parameters. We run the evolutionary process for 300 generations and finally select at the end of the evolutionary process the DSP rule that achieved the best fitness value.

\subsection{Hill Climbing}\label{sec:HCalgorithm}
We use the HC algorithm as a baseline to compare the results of the DSP. The HC algorithm, visualized in Figure~\ref{fig:hillClimbingFramework}, is a single-solution local search algorithm~\cite{de2006fundamentals}. It is analogous to the DSP, except the fact that it is used as a direct encoding optimization approach where domain knowledge (i.e. knowledge of the neuron activations) is not introduced into the optimization process. Therefore, it provides a suitable baseline to assess the effectiveness of the domain-based heuristic used in the DSP.

A trial of the HC algorithm aims to find an optimum set of RNN weights that allows the agent to reach a given goal position, starting from the starting position. All the connection weights of the RNN (644 in total) are directly encoded into a single real-valued vector that represents an $\boldsymbol{Agent}$ (candidate solution). At the beginning of the algorithm, each variable of the $\boldsymbol{Agent}$ is randomly initialized in the range $[-1,1]$ with uniform probability. The $\boldsymbol{Agent}$ is then evaluated and assigned as $\boldsymbol{Agent_{best}}$, with fitness equal to $EP_{best}$. After the evaluation, $\boldsymbol{Agent_{best}}$ is perturbed using a \textit{perturbation heuristic} to generate a new $\boldsymbol{Agent}$ as follows:
\begin{equation}
\boldsymbol{Agent} = \boldsymbol{Agent_{best}} + \sigma \cdot \mathcal{N} (0,1)
\label{eq:perturbationHeuristic}
\end{equation}
where $\mathcal{N}$ is a random vector whose length is the same as that of $\boldsymbol{Agent_{best}}$, and each dimension is independently sampled from a Gaussian distribution with zero mean and unitary standard deviation, scaled by the parameter $\sigma$.

The $\boldsymbol{Agent}$ is evaluated and if its $EP_e$ is smaller than $EP_{best}$, $\boldsymbol{Agent_{best}}$ is replaced by the new $\boldsymbol{Agent}$. The perturbation and evaluation processes are performed iteratively for 100 episodes (evaluations). The overall performance of the HC algorithm is then obtained by averaging the $EP_{best}$ from 40 trials (consisting of 5 trials for 8 distinct goal positions), that is the same evaluation procedure used to evaluate DSP rules.

The performance of the HC algorithm may depend on the parameter of the perturbation heuristic. Thus, to provide a fair comparison with DSP, we optimize the parameter $\sigma \in [0,1]$ with respect to the hyper-parameters of the RNN model $\alpha_h \in [0,1]$ and $\alpha_o \in [0,1]$ by using a GA with the same settings used to optimize the continuous part of the DSP rules (see Section~\ref{sec:GA}). We refer these three parameters as HC parameters.

\section{Results}\label{sec:results}

\begin{figure}[t]
\begin{subfigures}
\subfloat[]{\includegraphics[width=0.5\columnwidth]{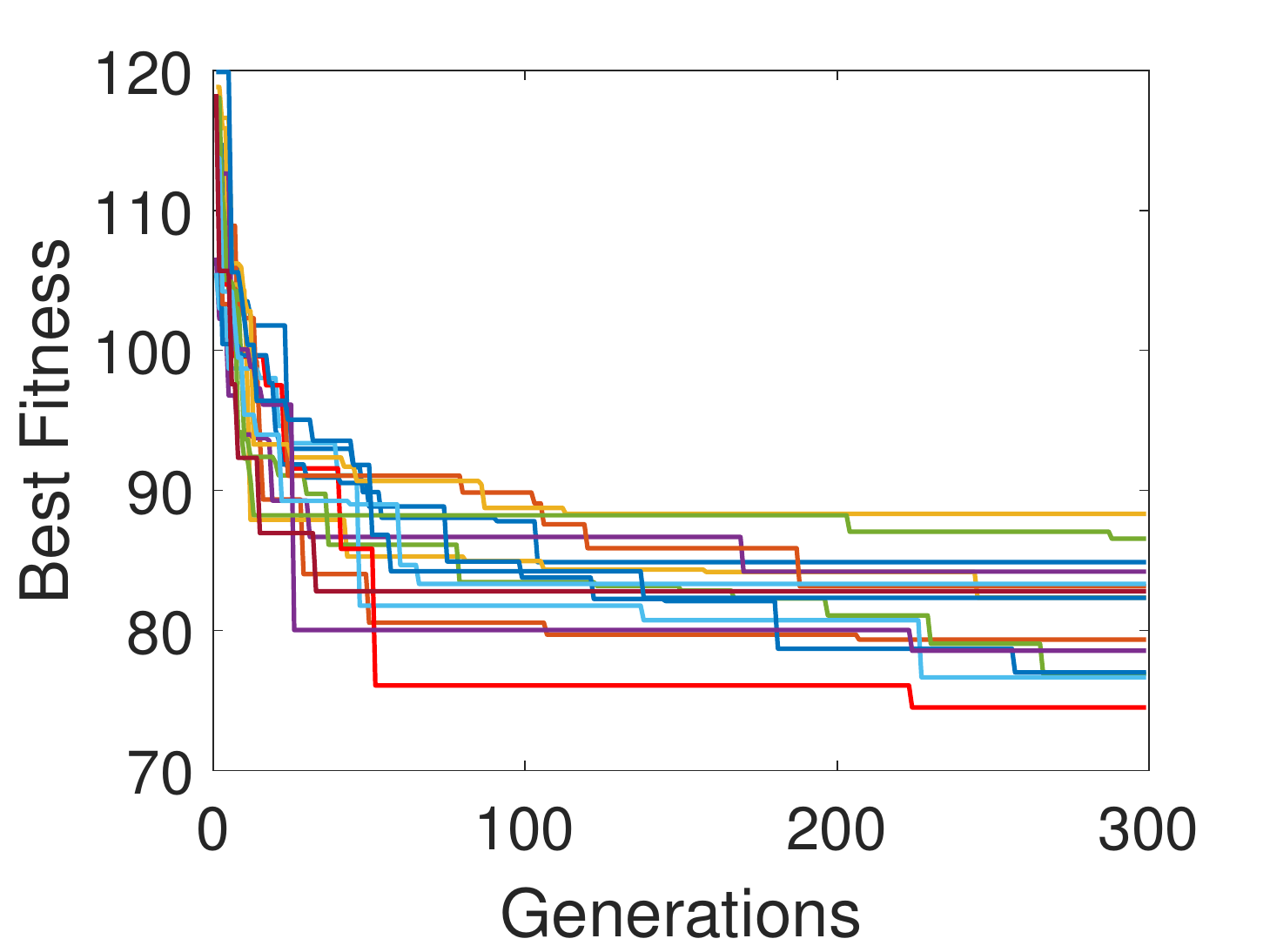}\label{fig:generationsDSP}} \subfloat[]{\includegraphics[width=0.5\columnwidth]{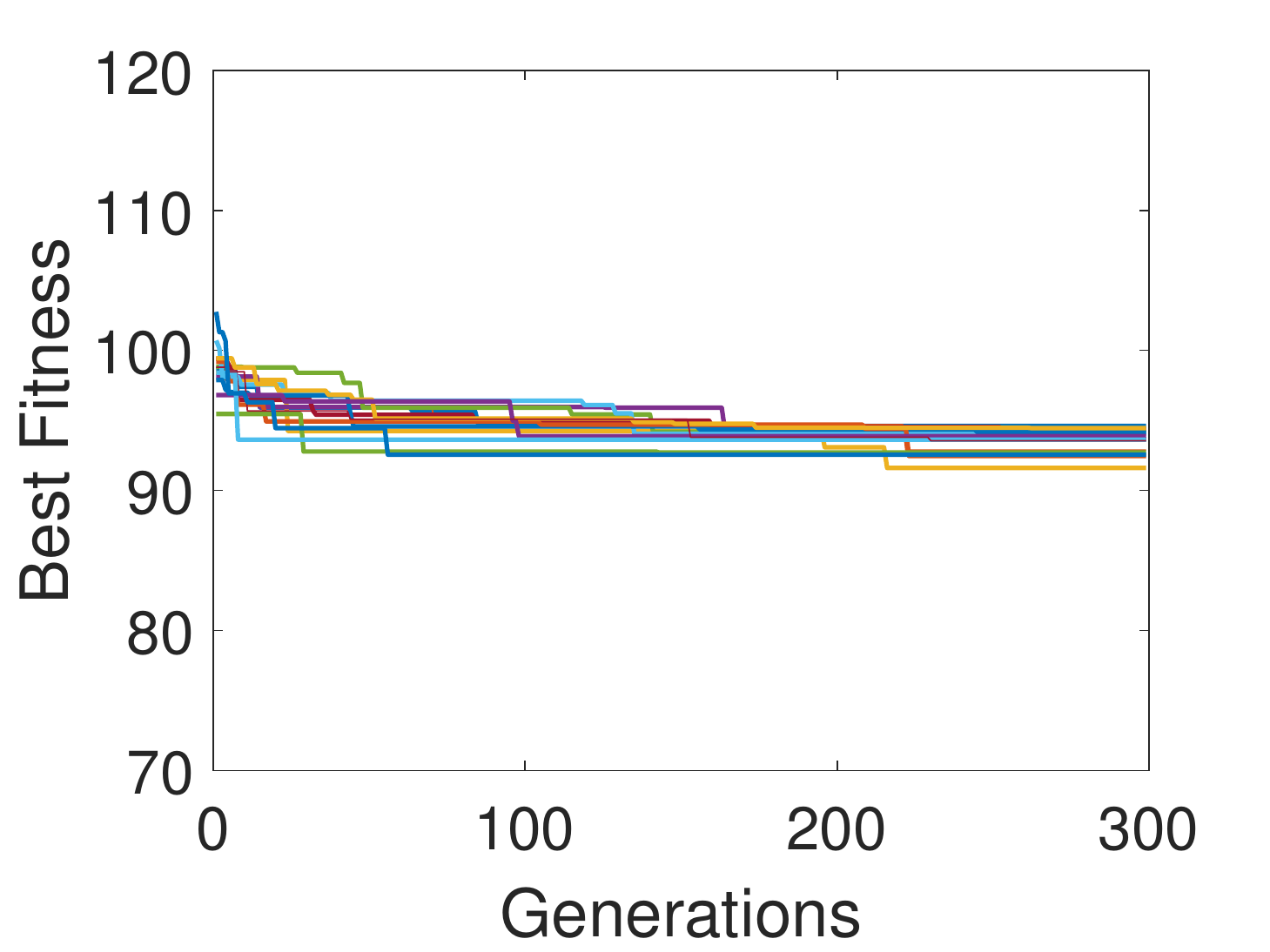}\label{fig:generationsHC}}
\end{subfigures}
\caption{The change of the best fitness during 15 independent evolutionary processes for optimizing (a) the DSP rules and (b) the HC parameters. The $y$-axes of figures are scaled between 70--120 to allow a better visual comparison.}\label{fig:generations}
\end{figure}

Figures~\ref{fig:generationsDSP} and \ref{fig:generationsHC} show the change of the best fitness value during 15 independent evolutionary optimization processes of the DSP rules and the HC parameters respectively. We run all the experiments on a single-core Intel Xeon E5 3.5GHz computer; therefore, we fix the number of generations to 300, to keep the runtime of the algorithm reasonably limited. In each generation, the best fitness value obtained by the DSP rules and the HC algorithm with evolved parameters are shown. A complete list of the evolved DSP rules can be found on an extended version of this paper online\footnote{An extended version of the paper with complete list of evolved DSP rules can be found here: https://arxiv.org/abs/1903.09393.}. 

The initial DSP rules obtain an average fitness of 113.79, with a standard deviation of 4.30; at the end of the evolutionary processes, they achieve an average fitness of 81.39, with a standard deviation of 4.02. On the other hand, the initial HC parameters obtain an average fitness value of 98.71, with a standard deviation of 1.64; and at the end of the evolutionary process, they achieve an average fitness of 93.50, with a standard deviation of 0.87. We used the Wilcoxon test to statistically assess the significance of the results produced by the evolutionary processes~\cite{wilcoxon1945}. The null-hypothesis that the mean of the results produced by two processes are the same is rejected if the $p$-value is smaller than $\alpha = 0.05$. In our case, the results of the evolutionary processes of DSP rules are statistically different (better than the HC results) with a $p$-value of $3.3\times 10^{-6}$.
\begin{figure}[t]
\begin{subfigures}
\subfloat[]{\includegraphics[width=0.5\columnwidth]{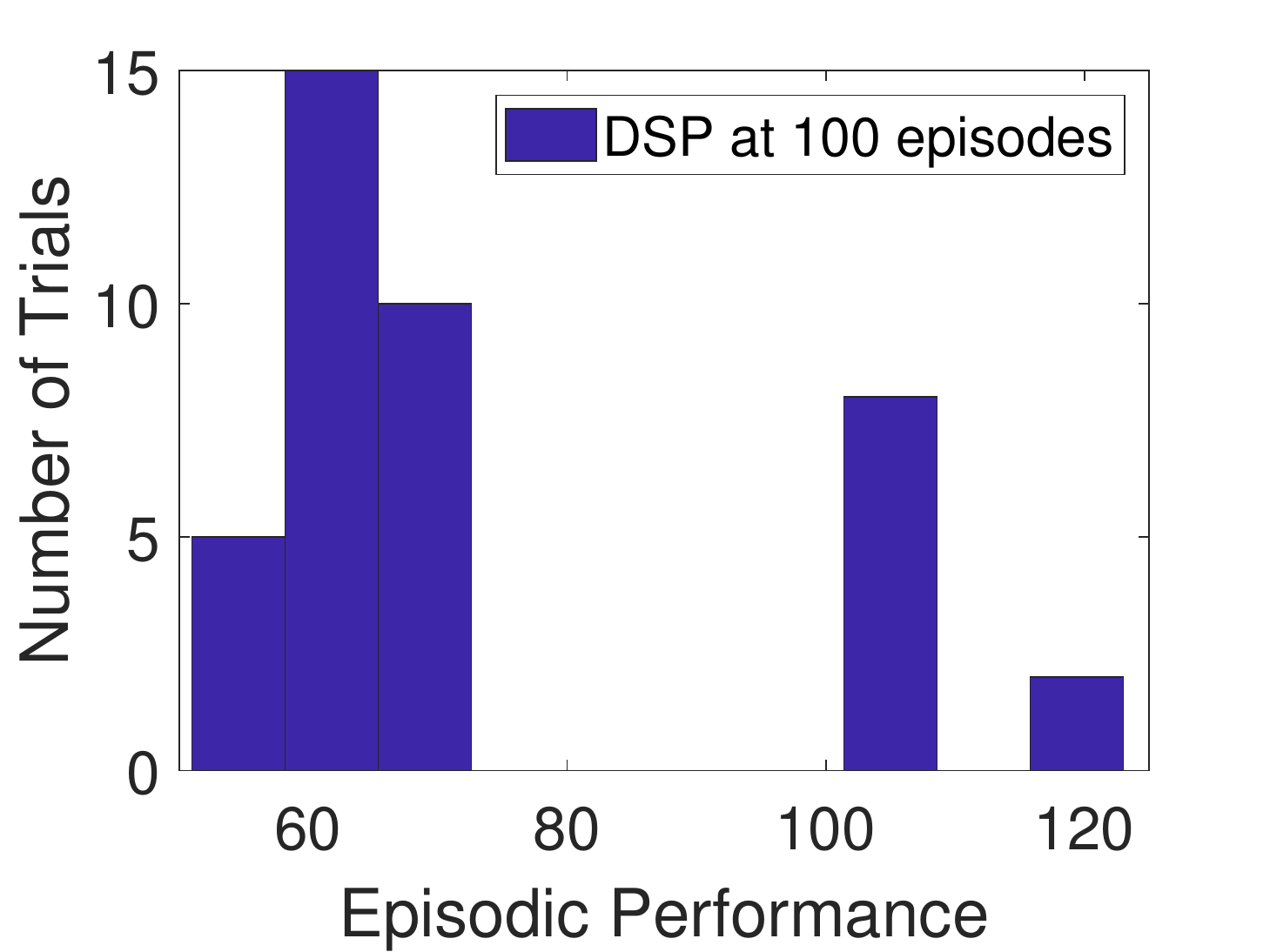}\label{fig:DSPdistribution100episodes}}
\subfloat[]{\includegraphics[width=0.5\columnwidth]{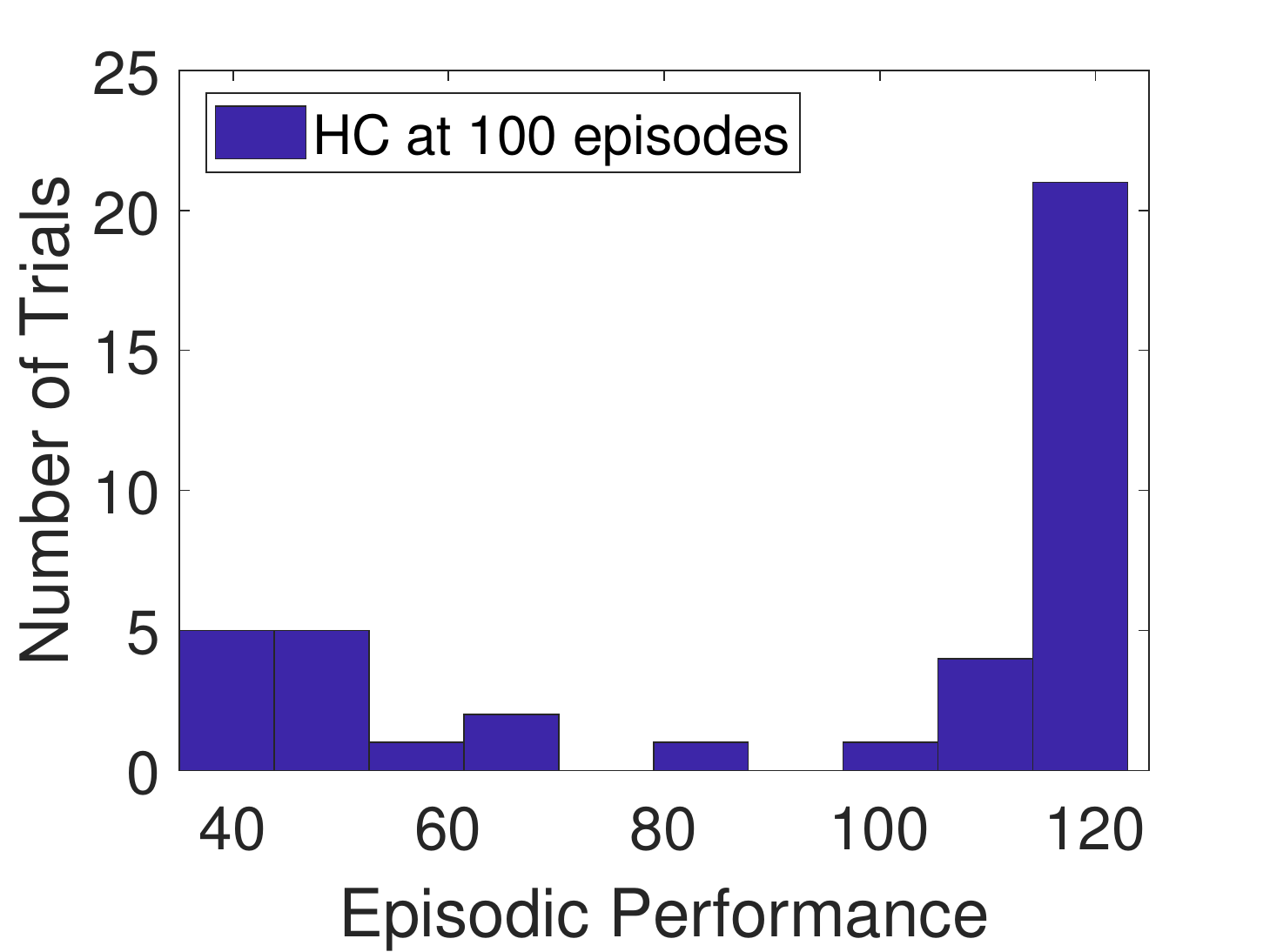}\label{fig:HCdistribution100episodes}}
\end{subfigures}
\caption{The distribution of the episodic performance of 40 trials using the best performing evolved DSP rule (a) and HC parameters (b) trained for 100 episodes.} \label{fig:distributionOfResutls100Episodes}
\end{figure}

\begin{figure}[t]
\begin{subfigures}
\subfloat[]{\includegraphics[width=0.5\columnwidth]{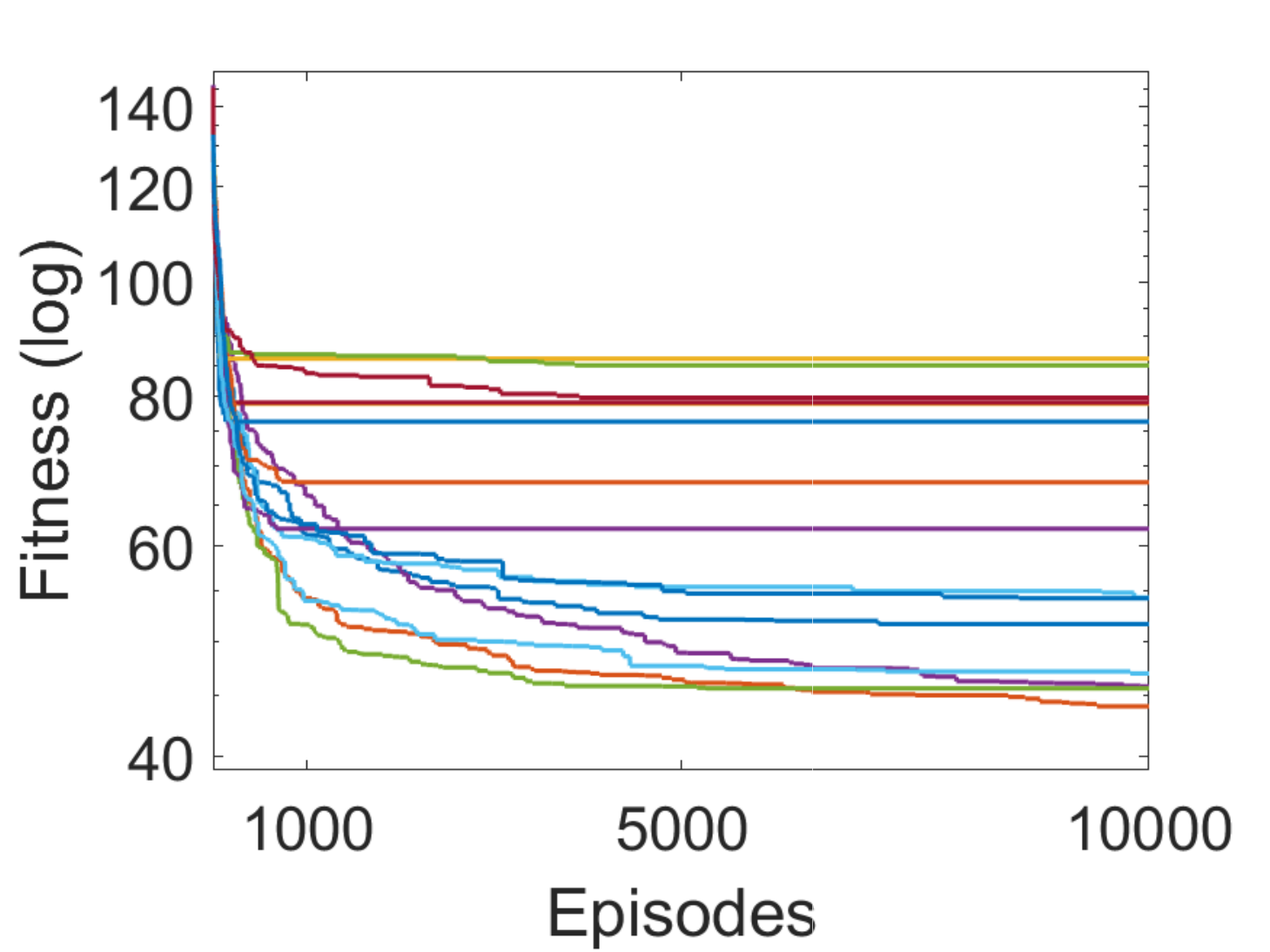}\label{fig:DSP10k}} \subfloat[]{\includegraphics[width=0.5\columnwidth]{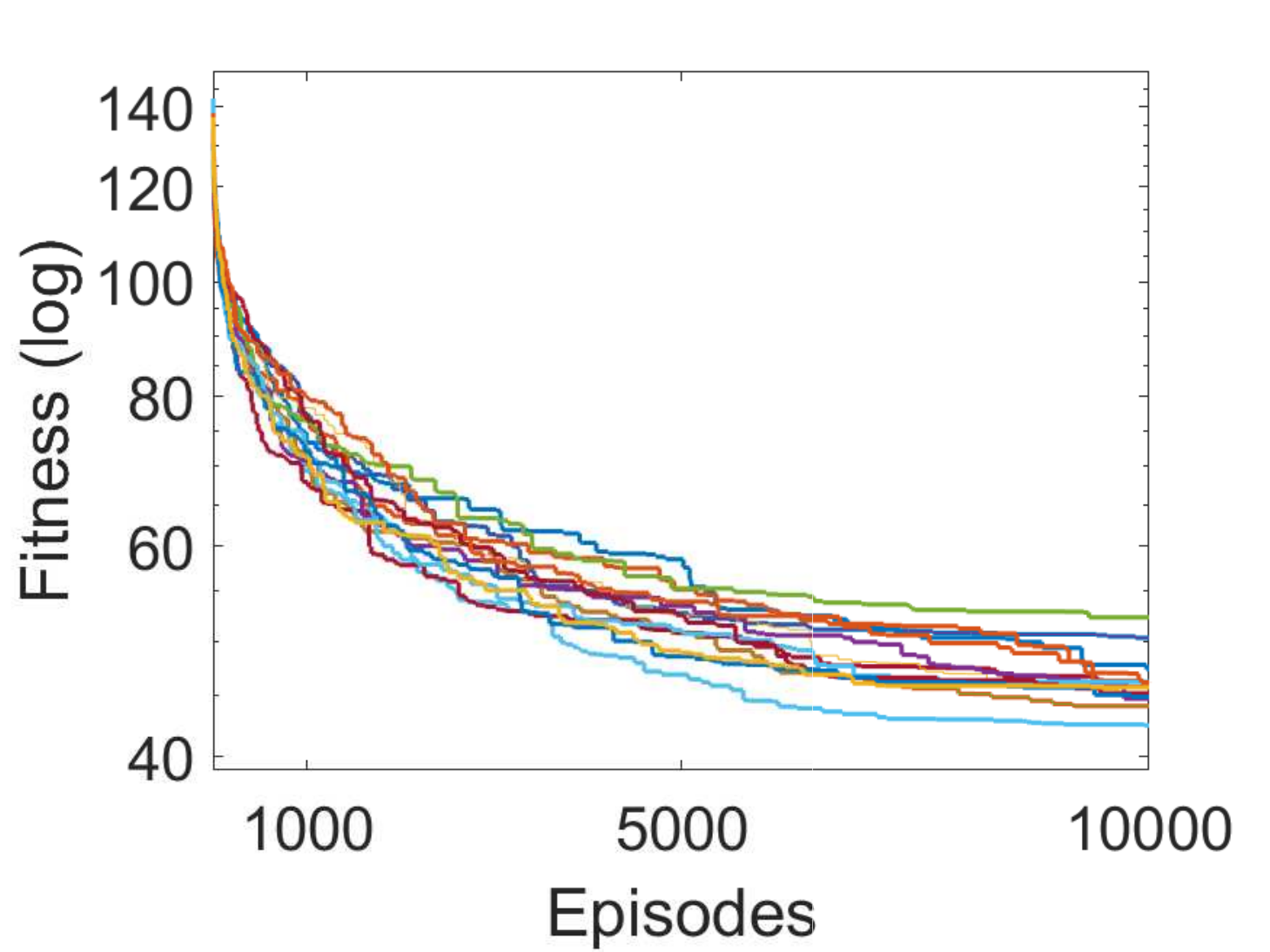}\label{fig:HC10k}}
\\
\subfloat[]{\includegraphics[width=0.5\columnwidth]{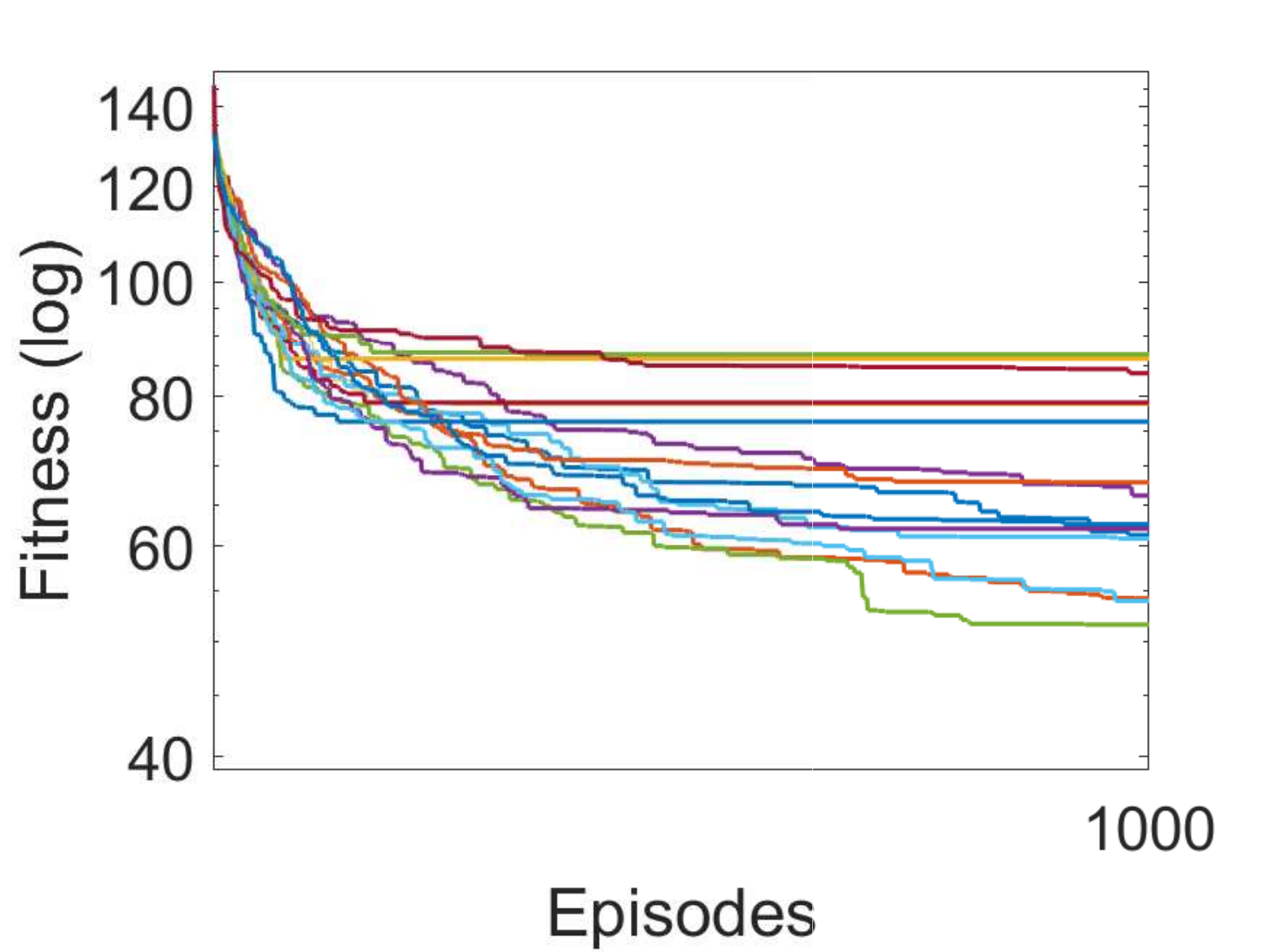}\label{fig:DSP1k}} 
\subfloat[]{\includegraphics[width=0.5\columnwidth]{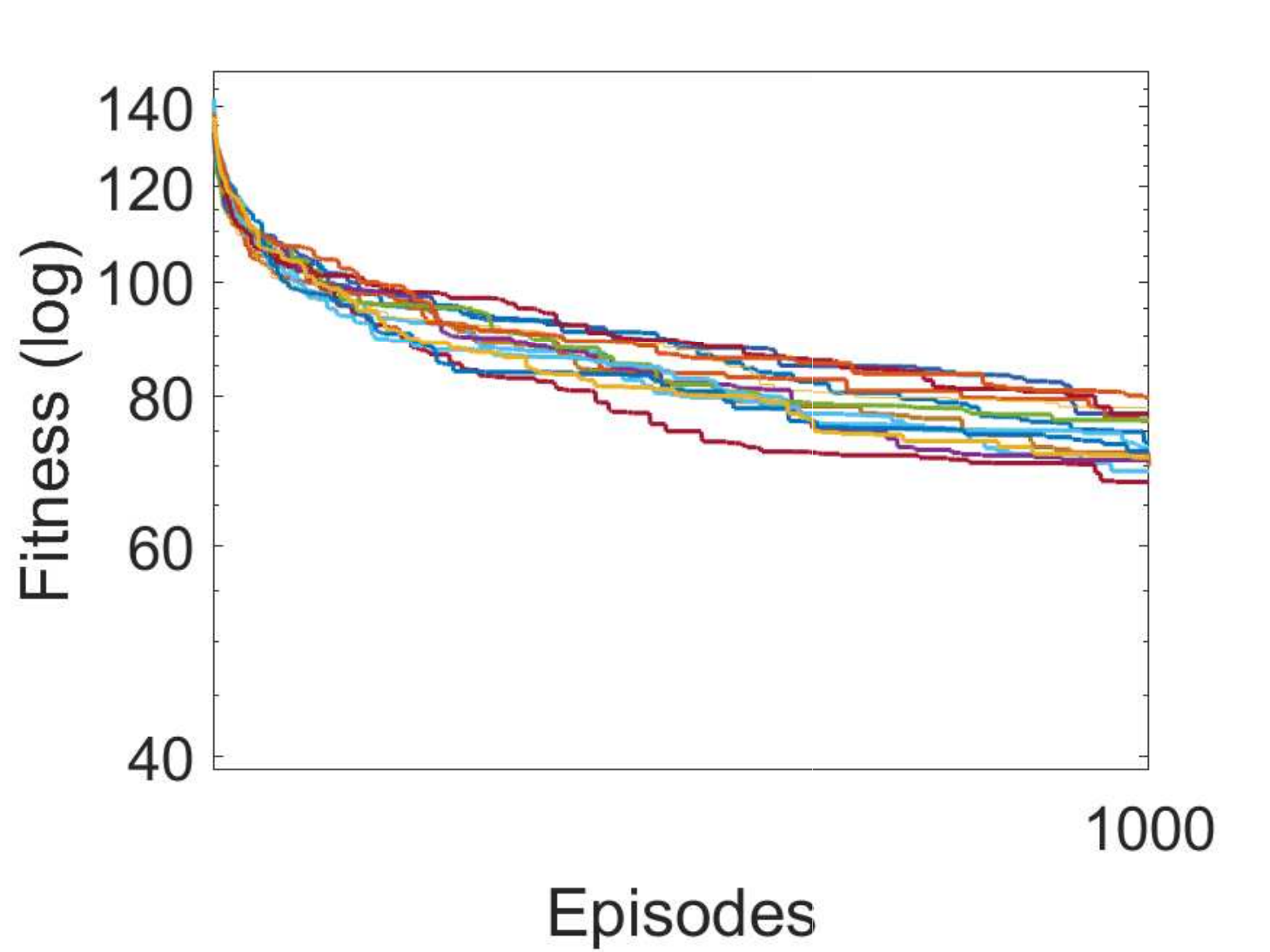}\label{fig:HC1k}}
\end{subfigures}
\caption{(a) The fitness values of the best evolved DSP rules, and (b) the fitness values of the best evolved HC parameters, both independently tested for 40 trials with 10000 episodes. Figures~\ref{fig:DSP1k} and \ref{fig:HC1k} are magnified views between 1--1000 episodes of (a) and (b) respectively.}\label{fig:resutlsOn10k}
\end{figure}

The distribution of the episodic performance of the best evolved DSP rule and HC parameters are given in Figure~\ref{fig:distributionOfResutls100Episodes}. The trials with an episodic performance smaller than 100 indicate that the goal is achieved in that trial. Thus, 75\% of the 40 trials reached the goal when the agents are trained with the DSP rules. On the other hand, only 35\% of the trials reached the goal when the agents are trained using the HC. A Wilcoxon rank-sum test (calculated on the EPs) shows that the DSP rule is better with a $p$-value of $0.03$. 
The results show that the training process with 100 episodes does not seem to be sufficient for the HC algorithm to provide results as good as the results provided by the DSP rules. Moreover, it may be possible to improve the success percentage of achieving the goal using the DSP rule by increasing the number of episodes. To test this, we separately tested the DSP rules and the HC with evolved parameters provided by the multiple runs of the GA on the same task settings with 10000 episodes. The results are given in Figure~\ref{fig:resutlsOn10k} where we show the change of the fitness values (average of 40 trials) w.r.t. the episode number during the training of the DSP rules and the HC. 

The best DSP rule achieved fitness of 54.27 and 44.10, and the HC with the best parameters achieved fitness of 69.35 and 42.5 in 1000 and 10000 episodes respectively. The results indicate that the DSP rules converge at a better fitness value faster than the HC. However, when the number of episodes is increased (for 10000), the HC achieves slightly (a fitness value difference of $1.6$) better performance than the best DSP rule. The $p$-values for the Wilcoxon tests for the results at episodes 1000 and 10000 are $0.02$ and $0.3$; thus, at 1000 episodes DSP rule is significantly better than HC, whereas at 10000 there is no significant difference between their results.
\begin{figure}[t]
\begin{subfigures}
\subfloat[]{\includegraphics[width=0.5\columnwidth]{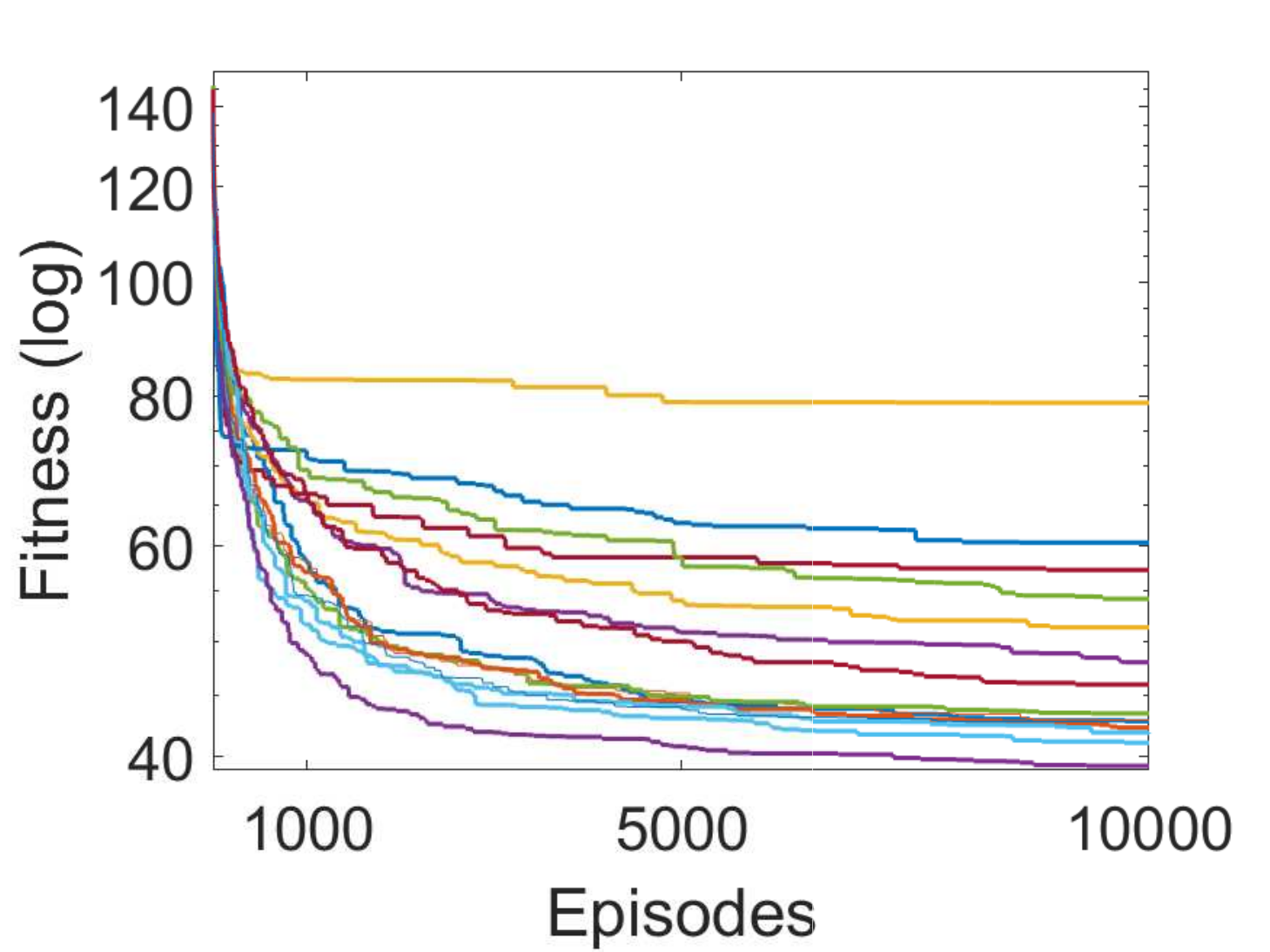}\label{fig:DSPIterated10k}}
\subfloat[]{\includegraphics[width=0.5\columnwidth]{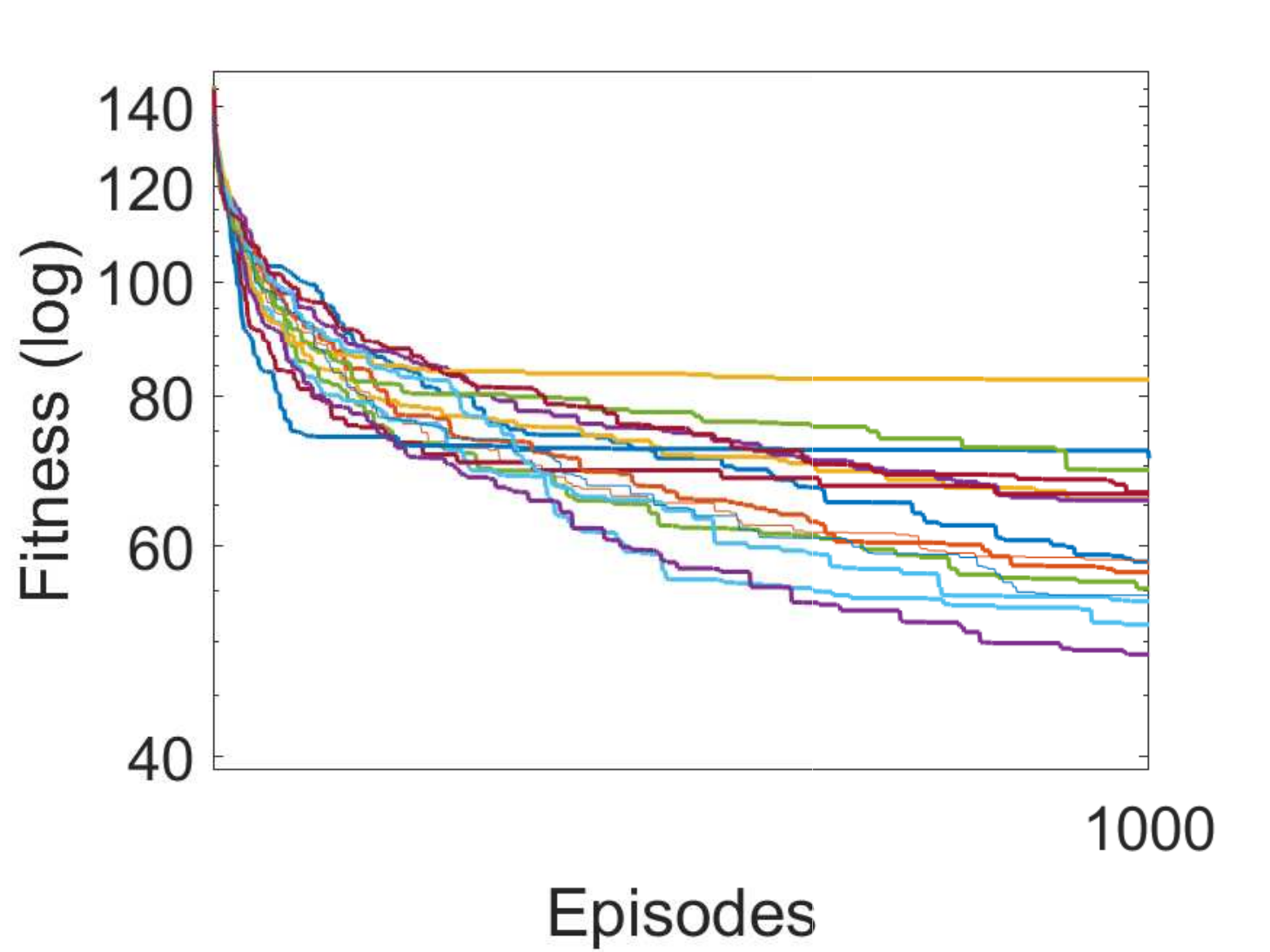}\label{fig:DSPIterated1k}}
\end{subfigures}
\caption{The fitness values of the best evolved delayed plasticity rules tested for 40 trials for 10000 episodes, with iterative re-sampling every 100 episodes. Figure~\ref{fig:DSPIterated1k} is a magnified view of the same results between episodes 1--1000.}\label{fig:DSPIterated}
\end{figure}

We observe that some of the DSP rules seem to get stuck at a local optimum after around 100 episodes, which may be due to the fact that they were optimized for 100 episodes. Thus, to reduce this effect we used an iterative re-sampling approach to randomly reset all the weights of the networks (from the initial domain) at every 100 episodes without resetting the best found fitness value. Thus, in the visualization, the fitness value does not appear worse than the best fitness due to each re-initialization. The training process generated by the iterative re-sampling based DSP rules is provided in Figure~\ref{fig:DSPIterated}. The best DSP rule achieves fitness values of 48.72 and 39.32 at 1000 and 10000 episodes. A perfect agent (an average of the distances of starting and 8 goal positions found by the A* algorithm) achieves a fitness value of 38.5. For completeness, we also performed additional experiments for the HC using the iterative re-sampling approach. However, we observed in this case that their results were not better than the standard HC with the tested settings. 
%
\begin{figure}[t]
\begin{subfigures}
\subfloat[]{\includegraphics[width=0.45\columnwidth]{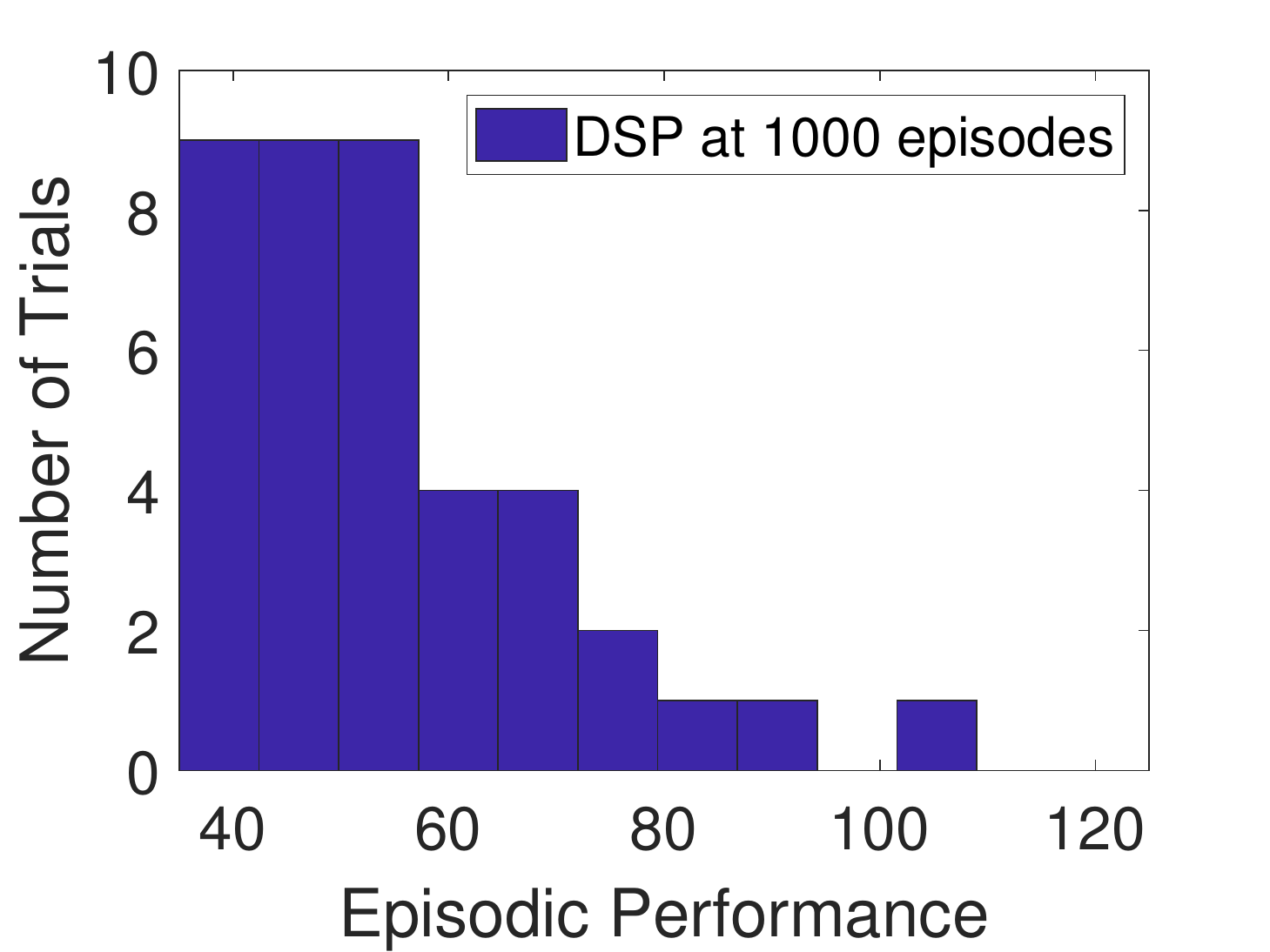}\label{fig:DSPdistribution1kEpisodes}}
\subfloat[]{\includegraphics[width=0.45\columnwidth]{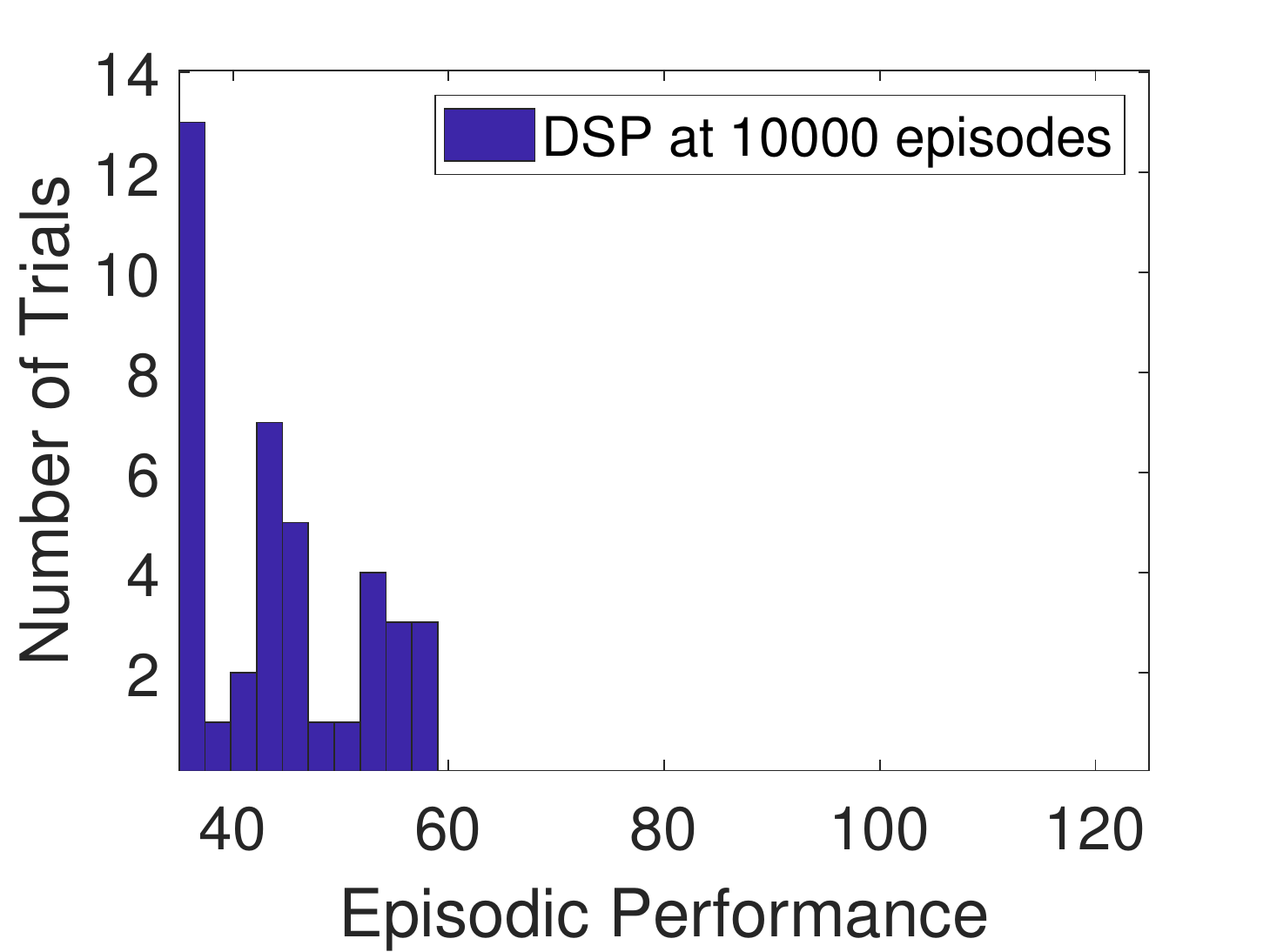}\label{fig:DSPdistribution10kEpisodes}}
\\
\subfloat[]{\includegraphics[width=0.45\columnwidth]{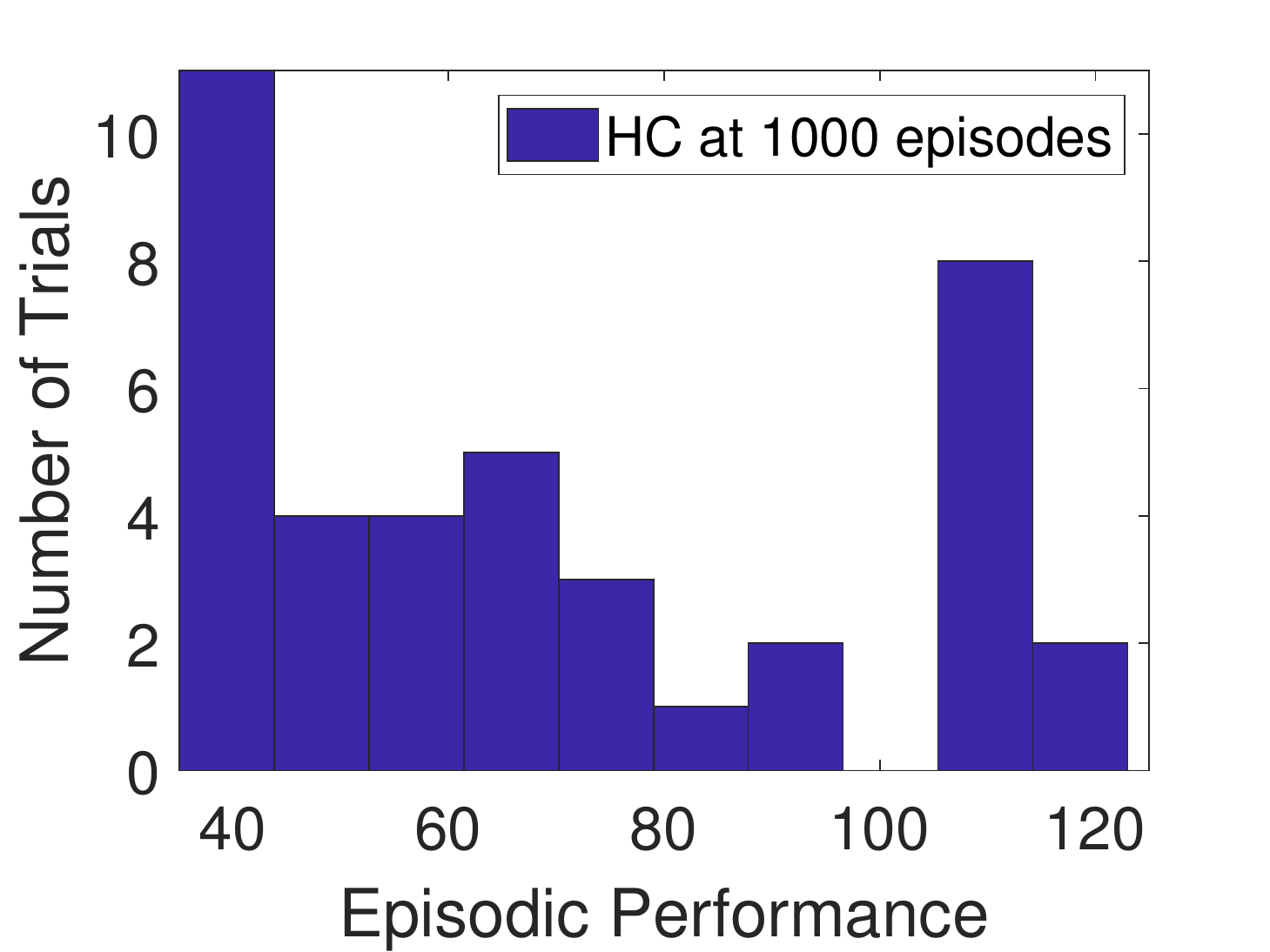}\label{fig:HCdistribution1kEpisodes}}
\subfloat[]{\includegraphics[width=0.45\columnwidth]{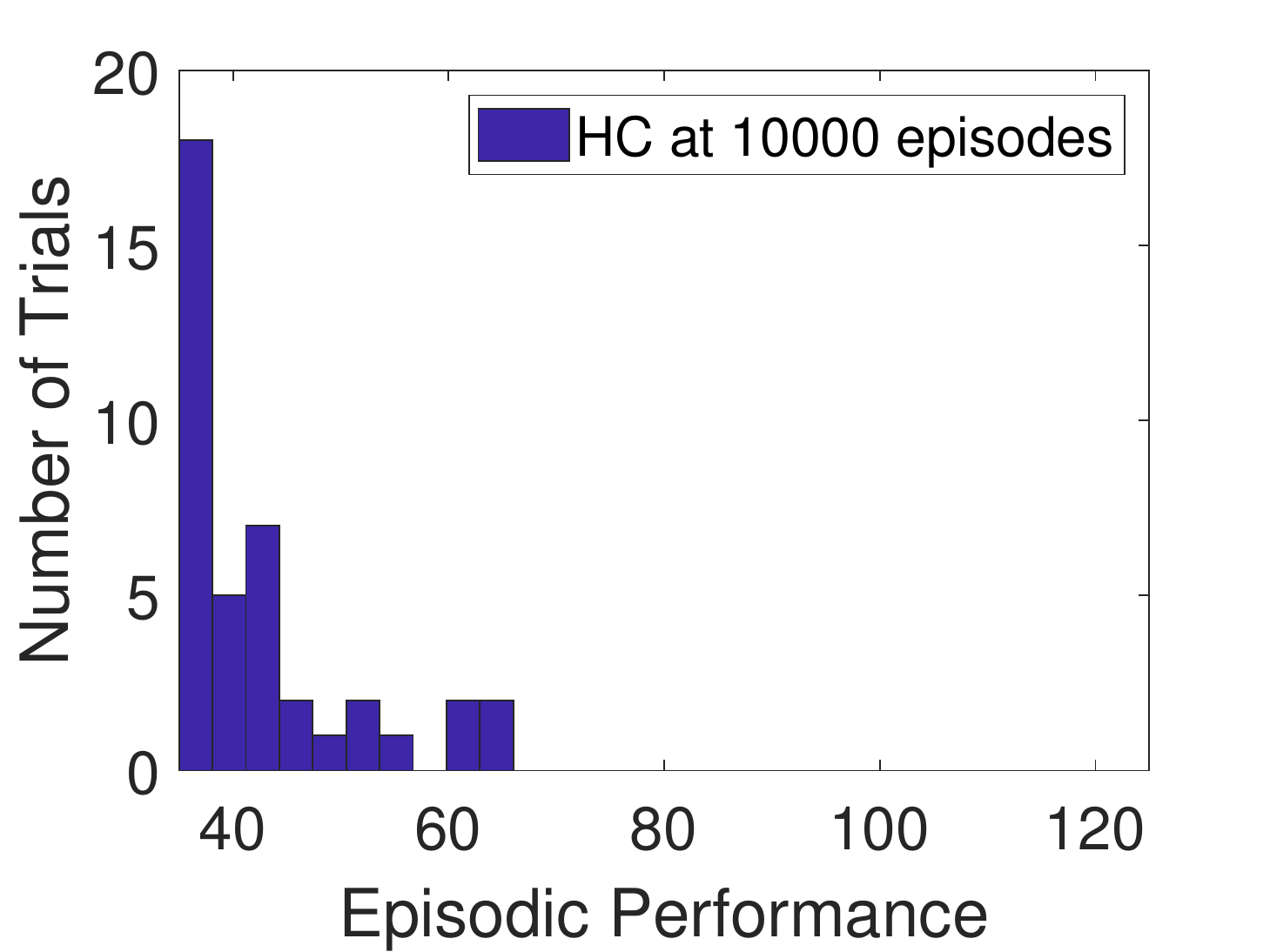}\label{fig:HCdistribution10kEpisodes}}
\\
\subfloat[]{\includegraphics[width=0.45\columnwidth]{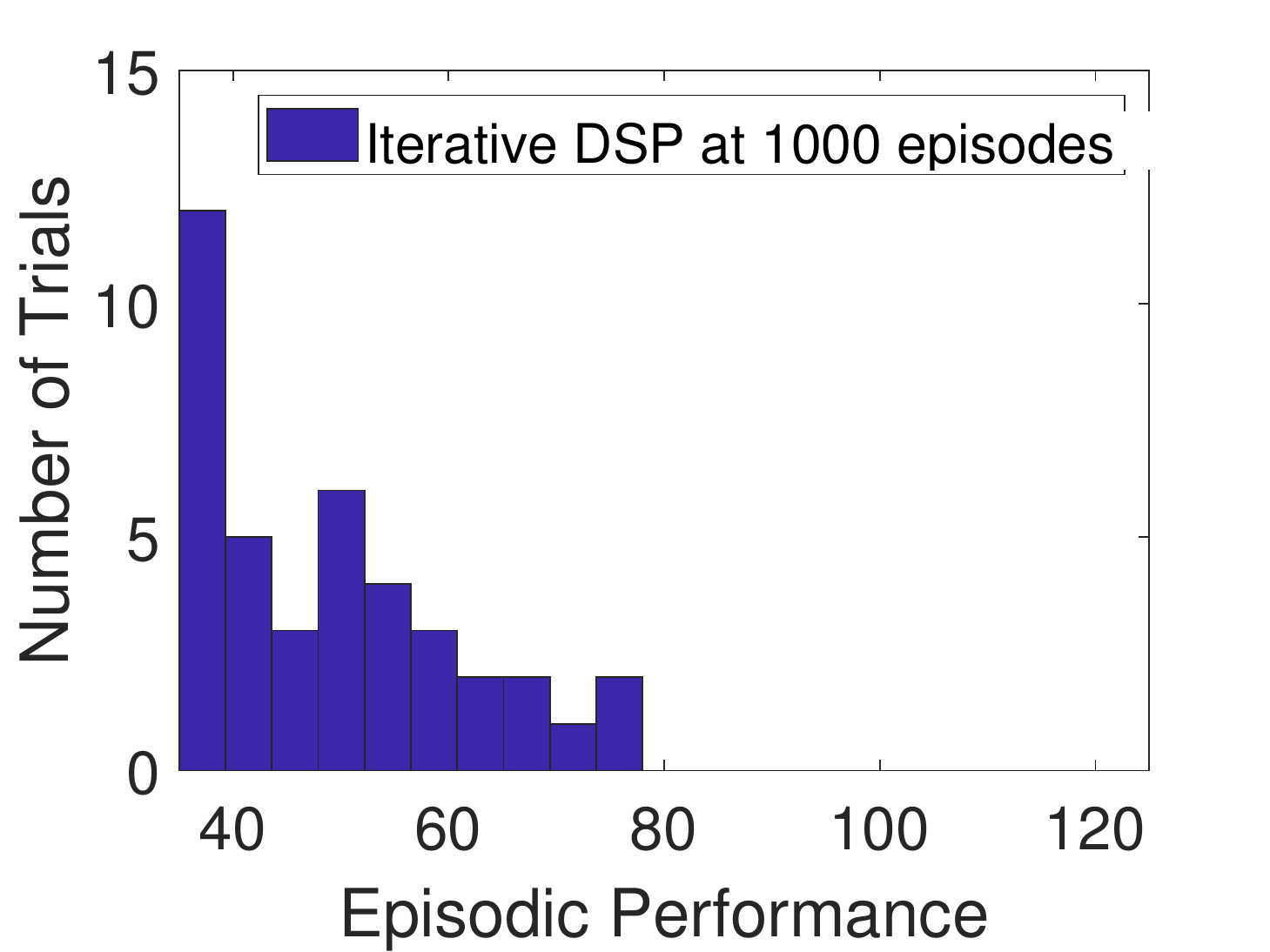}\label{fig:DSPdistribution1kEpisodesIterated}}
\subfloat[]{\includegraphics[width=0.45\columnwidth]{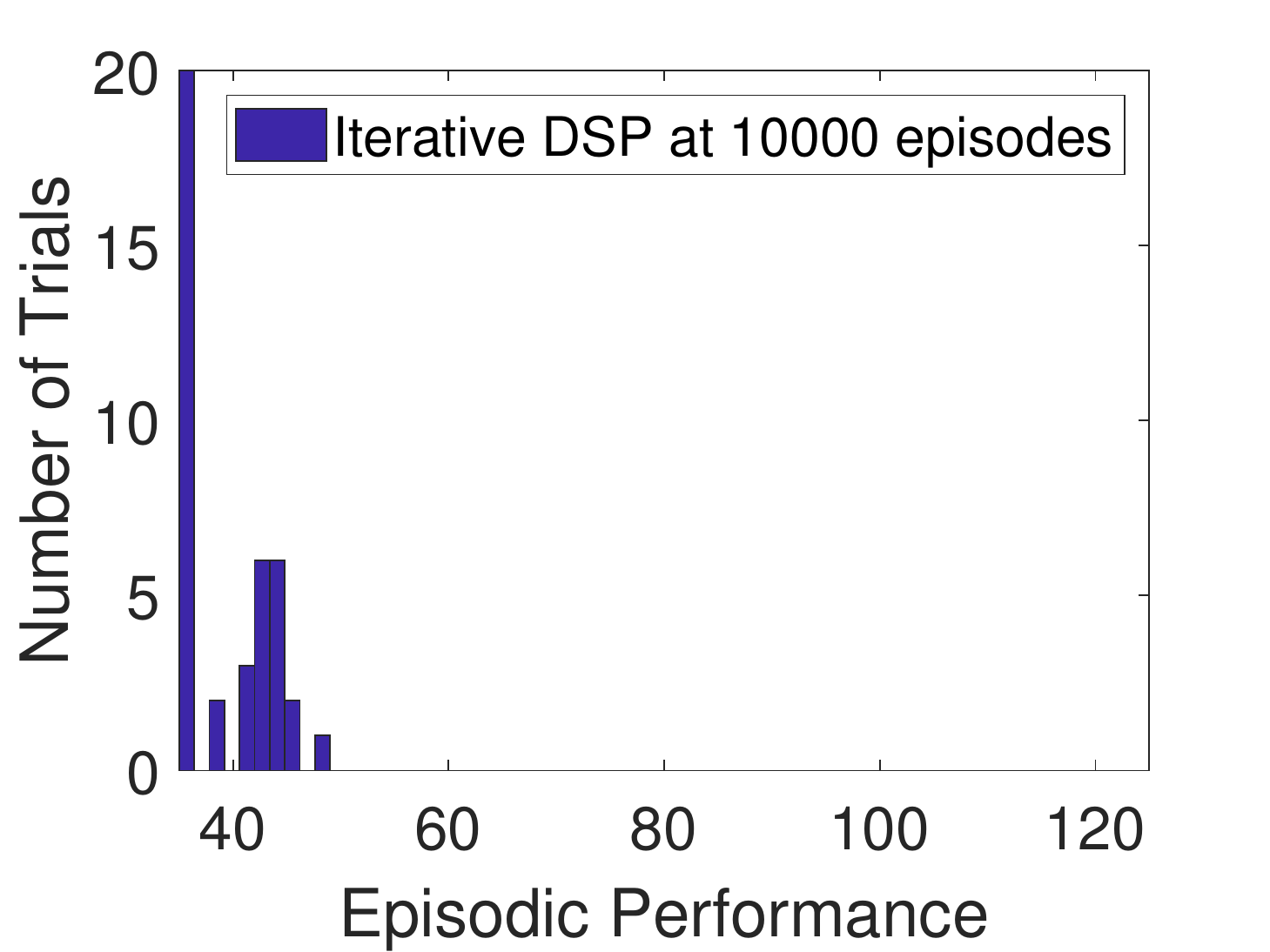}\label{fig:DSPdistribution10kEpisodesIterated}}
\end{subfigures}
\caption{The distribution of the episodic performance of 40 trials using the best DSP rule, HC, and DSP rule with iterative re-sampling approach at episodes 1000 and 10000 are given in (a) (b), (c) (d), and (e) (f) respectively.} \label{fig:distributionComparison10k}
\end{figure}

The distributions of the episodic performance of the best performing DSP rule, HC with best parameters and DSP with iterative approach at 1000 and 10000 episodes are given in Figure~\ref{fig:distributionComparison10k}. We fixed the minimum and maximum values of the $x$-axes of all figures to 35--125 for a better visual comparison. At 1000 episodes, all of the agents in 40 trials reach the goal with the smallest number of steps when the DSP rule with iterative re-sampling is used. On the other hand, 39 and 30 of the agents in 40 trials reach the goal when we use the DSP rule and HC respectively. At 10000 episodes, all of the agents in 40 trials reach the goal when the best heuristic from each approach is used. However, in the case of the DSP rule with iterative re-sampling, the agents reach the goal with the smallest number of steps on average.  
\begin{figure}[t]
\begin{center}
\includegraphics[width = \columnwidth]{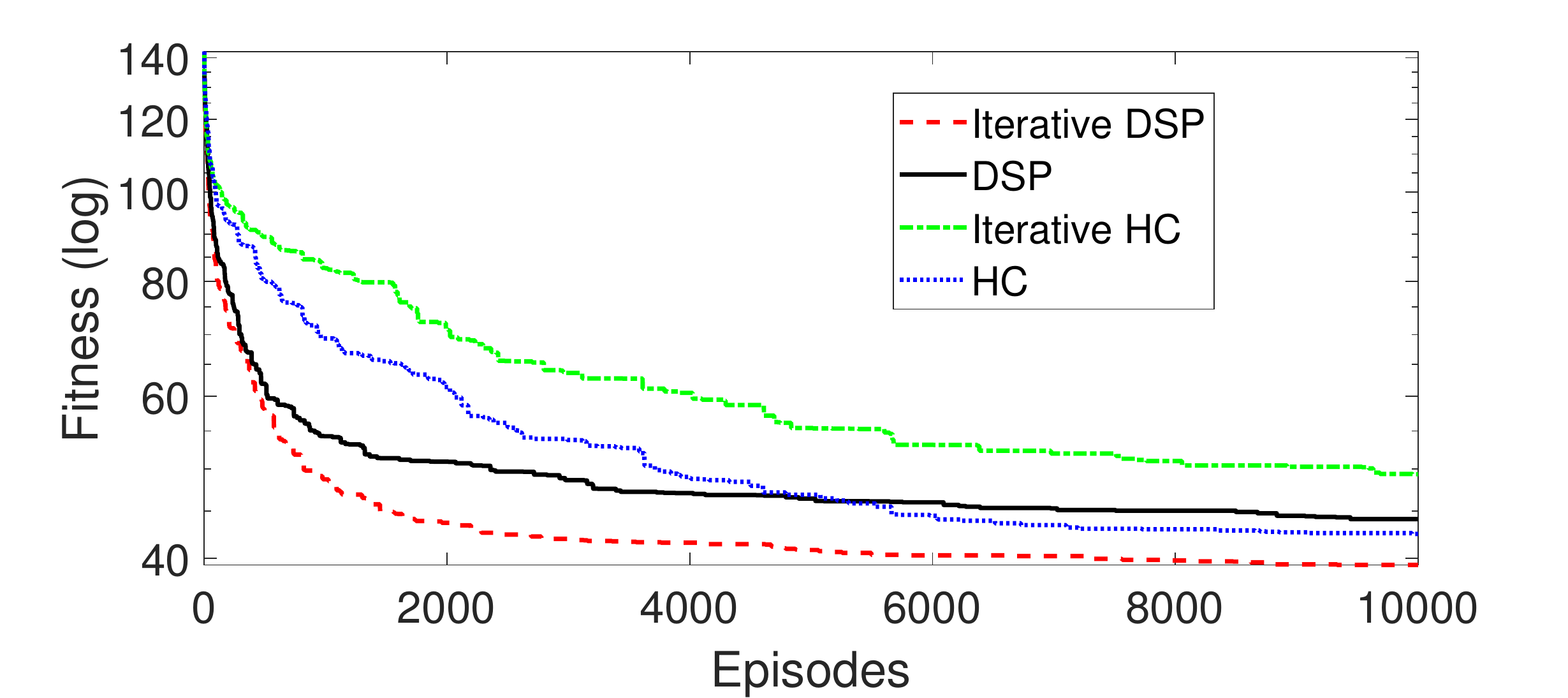}
\caption{The fitness of the best evolved DSP (standard and iterative re-sampling variants) rule and the HC with best evolved parameters (standard and iterative re-sampling variants) tested on 10000 episodes.}
\label{fig:comparison10k}
\end{center}
\end{figure}
Figure~\ref{fig:comparison10k} shows the comparison of the best evolved DSP\footnote{ We have recorded the performance of the agents on the triple T-maze task after training with the best evolved DSP rule for 100, 1000 and 10000 episodes. A video of the experiment is available online at: \url{https://youtu.be/J0WYMrAMSdU}.}, iterative DSP, HC and iterative HC heuristics with best evolved parameters. Iterative HC performs worse than the HC. On the other hand, the HC was able to outperform the DSP at around 5000 generations, while it could not perform better than the iterative DSP.

\section{Conclusions} \label{sec:conclusion}

When the reinforcement signals are available after a certain period of time, it may not be possible to associate the activations of the neurons during this period with the reinforcement signals, and perform synaptic updates using Hebbian learning. In this work, we proposed the NATs, i.e. additional data storage in each synapse to keep track of neuron activations. We used DSP rules that take into account the NATs and delayed reinforcement signals to perform synaptic updates. We used relative reinforcement signals that were provided after an episode based on the relative performance of the agent in a previous episode. Since the NATs introduce knowledge of neuron activations into the DSP rules, we compared their results to an analogous HC algorithm that performs random synaptic updates without any knowledge of the network activations.

We observed that the DSP rules were highly efficient at training networks with a smaller number of episodes compared to the HC, as they converge quickly at a better fitness than the HC. When they were tested on a larger number of episodes on the other hand, they seemed to be outperformed slightly by HC. We hypothesized that this could be due to the fact that the DSP rules were optimized for a relatively small number of episodes (100). When we tried a iterative re-sampling approach, the DSP rules provided the best results. On the other hand, the DSP rules introduce an additional complexity that requires storing/updating four parameters per synapse during the network computation, and looking-up the update rule based on the activation patterns for each synaptic update.  

We aim to apply the DSP on different ANN network models (i.e. continuous neuron activations) with various sizes, and test it on tasks with various complexity. It would also be interesting to investigate the adaptation capabilities of the networks with DSP when the environmental conditions change.

\section{Acknowledgements}
\noindent
      \begin{tabular}{p{0.15\linewidth} p{0.8\linewidth}}
      \raisebox{-0.8cm}{\includegraphics[height=.95cm]{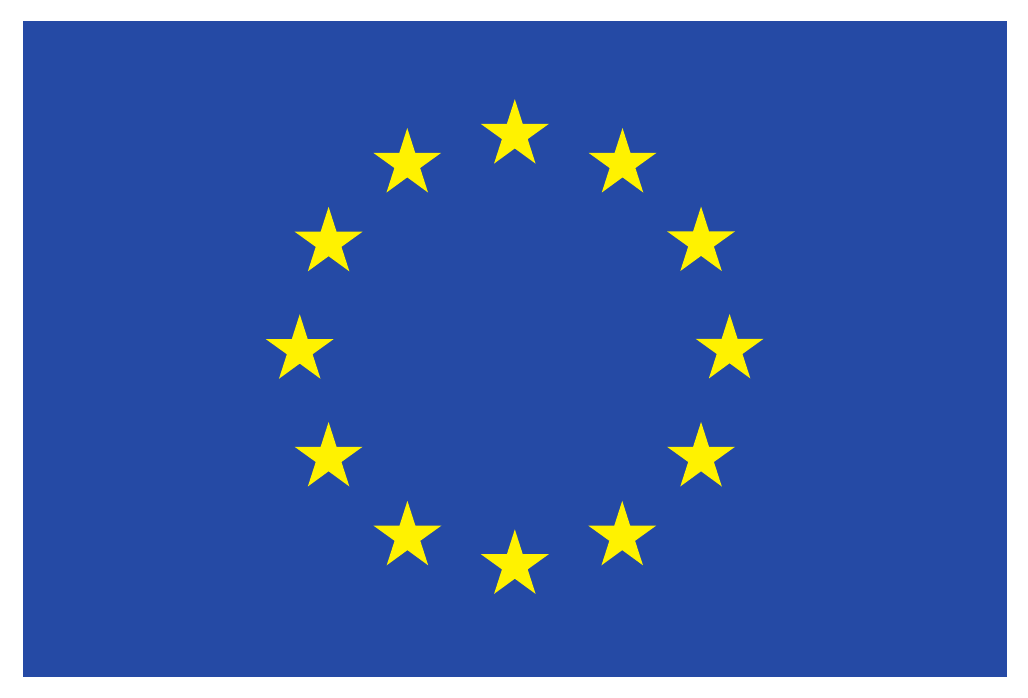}}
      &
  	{\small\fontfamily{arial}\selectfont This project has received funding from the European Union's Horizon 2020 research and innovation programme under grant agreement No: 665347.}
  	\end{tabular}


\onecolumn
\appendix
\section{Extended Results} \label{appx:extendedResults}
\subsection{Evolved DSP rules}
\begin{table}[ht]
\small
\caption{A complete list of the continuous parts of the 15 distinct evolved DSP rules and their fitness values after 10000 episodes. Their discrete parts can be found in Table \ref{tab:bestEvolvedDSPrule}.}\label{tab:continuousPartOfDSPs}
\begin{center}
\begin{tabular}{|c|c|c|c|c|c|c|c|c|c|c|c|c|c|c|c|}
\hline
\textbf{RuleID}	&	\textbf{1}	&	\textbf{2}	&	\textbf{3}	&	\textbf{4}	&	\textbf{5}	&	\textbf{6}	&	\textbf{7}	&	\textbf{8}	&	\textbf{9}	&	\textbf{10}	&	\textbf{11}	&	\textbf{12}	&	\textbf{13}	&	\textbf{14}	&	\textbf{15} \\ \hline	
$\boldsymbol{\eta}$	&	0.0317	&	0.0754	&	0.0720	&	0.0470	&	0.0302	&	0.0152	&	0.0422	&	0.0927	&	0.0569	&	0.2396	&	0.2082	&	0.4536	&	0.0507	&	0.2315	&	0.9619	\\ \hline
$\boldsymbol{\theta}$	&	0.2080	&	0.5574	&	0.3530	&	0.2763	&	0.1923	&	0.5547	&	0.5492	&	0.1277	&	0.8238	&	0.0534	&	0.2351	&	0.3967	&	0.6040	&	0.3909	&	0.3672	\\ \hline
$\boldsymbol{\alpha_h}$	&	0.1931	&	0.1654	&	0.1523	&	0.2253	&	0.0985	&	0.0454	&	0.2291	&	0.1319	&	0.2833	&	0.0947	&	0.1272	&	0.4958	&	0.1334	&	0.1859	&	0.5027	\\ \hline
$\boldsymbol{\alpha_o}$	&	0.2376	&	0.2255	&	0.6770	&	0.0214	&	0.0445	&	0.1633	&	0.0626	&	0.4402	&	0.2538	&	0.0947	&	0.2613	&	0.1028	&	0.4862	&	0.2039	&	0.5758	\\ \hline
\textbf{Fitness}	&	44.10	&	45.65	&	45.83	&	46.98	&	51.65	&	54.28	&	54.35	&	62.05	&	67.85	&	76.25	&	78.70	&	79.10	&	79.85	&	84.98	&	86.08	\\ \hline

\end{tabular}
\end{center}
\end{table}

\begin{table}[ht]
\small
\caption{A complete list of 15 distinct evolved DSP rules given by the columns $\Delta w_1$ through $\Delta w_{15}$. Their continuous parts can be found in Table \ref{tab:continuousPartOfDSPs}. First four columns specify neuron activation traces where the first and second bits represent activations of pre- and post-synaptic neurons (i.e. $00$ is when pre- and post-synaptic neurons are in a non-active state.), and the fifth column specify the modulatory signal $m$.}\label{tab:bestEvolvedDSPrule}
\begin{tabular}{|c|c|c|c|c|c|c|c|c|c|c|c|c|c|c|c|c|c|c|c|}
\hline
\multicolumn{4}{|c|}{$\boldsymbol{NAT^{\theta}}$}                                                                 & \multirow{2}{*}{$\boldsymbol{m}$} & \multirow{2}{*}{$\boldsymbol{\Delta w_1}$}  & \multirow{2}{*}{$\boldsymbol{\Delta w_2}$} & \multirow{2}{*}{$\boldsymbol{\Delta w_3}$} & \multirow{2}{*}{$\boldsymbol{\Delta w_4}$} & \multirow{2}{*}{$\boldsymbol{\Delta w_5}$} & \multirow{2}{*}{$\boldsymbol{\Delta w_6}$} & \multirow{2}{*}{$\boldsymbol{\Delta w_7}$} & \multirow{2}{*}{$\boldsymbol{\Delta w_8}$} & \multirow{2}{*}{$\boldsymbol{\Delta w_9}$} & \multirow{2}{*}{$\boldsymbol{\Delta w_{10}}$} & \multirow{2}{*}{$\boldsymbol{\Delta w_{11}}$} & \multirow{2}{*}{$\boldsymbol{\Delta w_{12}}$} & \multirow{2}{*}{$\boldsymbol{\Delta w_{13}}$} & \multirow{2}{*}{$\boldsymbol{\Delta w_{14}}$} & \multirow{2}{*}{$\boldsymbol{\Delta w_{15}}$} \\ \cline{1-4}
$\boldsymbol{00}$             & $\boldsymbol{01}$            & $\boldsymbol{10}$           & $\boldsymbol{11}$            &                             &  & &&&&&&&&&&&&&                 \\ \hline\hline

0	&	0	&	0	&	0	&	-1	&	1	&	1	&	0	&	-1	&	-1	&	1	&	1	&	-1	&	0	&	1	&	1	&	0	&	0	&	0	&	-1	\\ \hline
0	&	0	&	0	&	0	&	1	&	1	&	1	&	1	&	0	&	-1	&	0	&	1	&	1	&	1	&	-1	&	1	&	-1	&	0	&	0	&	1	\\ \hline
0	&	0	&	0	&	1	&	-1	&	0	&	-1	&	-1	&	0	&	-1	&	-1	&	-1	&	-1	&	0	&	0	&	1	&	0	&	0	&	0	&	0	\\ \hline
0	&	0	&	0	&	1	&	1	&	-1	&	-1	&	-1	&	-1	&	0	&	-1	&	-1	&	-1	&	-1	&	-1	&	-1	&	-1	&	-1	&	-1	&	-1	\\ \hline
0	&	0	&	1	&	0	&	-1	&	1	&	0	&	1	&	0	&	0	&	0	&	0	&	-1	&	1	&	1	&	0	&	1	&	1	&	1	&	0	\\ \hline
0	&	0	&	1	&	0	&	1	&	1	&	1	&	1	&	1	&	1	&	1	&	1	&	1	&	1	&	0	&	1	&	1	&	1	&	0	&	1	\\ \hline
0	&	0	&	1	&	1	&	-1	&	0	&	0	&	0	&	-1	&	1	&	0	&	1	&	0	&	0	&	1	&	0	&	0	&	1	&	1	&	0	\\ \hline
0	&	0	&	1	&	1	&	1	&	-1	&	1	&	-1	&	0	&	-1	&	-1	&	1	&	1	&	0	&	-1	&	0	&	1	&	0	&	-1	&	-1	\\ \hline
0	&	1	&	0	&	0	&	-1	&	0	&	-1	&	0	&	1	&	-1	&	-1	&	-1	&	0	&	0	&	-1	&	-1	&	1	&	-1	&	0	&	0	\\ \hline
0	&	1	&	0	&	0	&	1	&	-1	&	-1	&	-1	&	-1	&	0	&	0	&	0	&	0	&	-1	&	-1	&	0	&	1	&	-1	&	0	&	0	\\ \hline
0	&	1	&	0	&	1	&	-1	&	0	&	0	&	0	&	1	&	0	&	1	&	-1	&	0	&	1	&	1	&	0	&	0	&	0	&	1	&	1	\\ \hline
0	&	1	&	0	&	1	&	1	&	1	&	-1	&	1	&	1	&	1	&	-1	&	1	&	0	&	-1	&	1	&	1	&	1	&	1	&	-1	&	1	\\ \hline
0	&	1	&	1	&	0	&	-1	&	1	&	0	&	1	&	1	&	1	&	0	&	0	&	-1	&	1	&	1	&	0	&	0	&	0	&	0	&	-1	\\ \hline
0	&	1	&	1	&	0	&	1	&	1	&	-1	&	-1	&	1	&	-1	&	0	&	1	&	-1	&	1	&	1	&	1	&	-1	&	0	&	0	&	1	\\ \hline
0	&	1	&	1	&	1	&	-1	&	0	&	1	&	1	&	1	&	-1	&	0	&	0	&	-1	&	1	&	-1	&	-1	&	0	&	0	&	0	&	0	\\ \hline
0	&	1	&	1	&	1	&	1	&	-1	&	0	&	1	&	1	&	0	&	0	&	-1	&	0	&	0	&	-1	&	0	&	0	&	1	&	1	&	0	\\ \hline
1	&	0	&	0	&	0	&	-1	&	0	&	1	&	1	&	-1	&	0	&	0	&	0	&	1	&	-1	&	-1	&	-1	&	-1	&	0	&	0	&	1	\\ \hline
1	&	0	&	0	&	0	&	1	&	0	&	0	&	0	&	1	&	0	&	0	&	0	&	0	&	0	&	0	&	0	&	0	&	1	&	0	&	0	\\ \hline
1	&	0	&	0	&	1	&	-1	&	0	&	-1	&	0	&	1	&	-1	&	-1	&	-1	&	-1	&	1	&	-1	&	-1	&	0	&	0	&	0	&	0	\\ \hline
1	&	0	&	0	&	1	&	1	&	1	&	-1	&	0	&	1	&	0	&	-1	&	1	&	0	&	0	&	-1	&	0	&	0	&	-1	&	-1	&	-1	\\ \hline
1	&	0	&	1	&	0	&	-1	&	0	&	-1	&	-1	&	-1	&	0	&	0	&	-1	&	1	&	0	&	-1	&	0	&	-1	&	0	&	0	&	1	\\ \hline
1	&	0	&	1	&	0	&	1	&	0	&	1	&	-1	&	0	&	1	&	1	&	1	&	1	&	-1	&	1	&	1	&	1	&	0	&	1	&	0	\\ \hline
1	&	0	&	1	&	1	&	-1	&	0	&	-1	&	0	&	1	&	-1	&	-1	&	1	&	-1	&	1	&	0	&	1	&	1	&	0	&	1	&	-1	\\ \hline
1	&	0	&	1	&	1	&	1	&	0	&	-1	&	-1	&	-1	&	1	&	-1	&	1	&	1	&	0	&	0	&	1	&	1	&	0	&	-1	&	-1	\\ \hline
1	&	1	&	0	&	0	&	-1	&	1	&	0	&	0	&	1	&	1	&	0	&	1	&	0	&	-1	&	-1	&	-1	&	-1	&	0	&	-1	&	-1	\\ \hline
1	&	1	&	0	&	0	&	1	&	0	&	0	&	1	&	-1	&	-1	&	0	&	-1	&	1	&	-1	&	-1	&	0	&	-1	&	-1	&	0	&	1	\\ \hline
1	&	1	&	0	&	1	&	-1	&	0	&	1	&	-1	&	-1	&	-1	&	0	&	-1	&	1	&	0	&	1	&	-1	&	1	&	0	&	1	&	1	\\ \hline
1	&	1	&	0	&	1	&	1	&	1	&	1	&	1	&	-1	&	1	&	0	&	0	&	-1	&	0	&	0	&	0	&	-1	&	0	&	0	&	1	\\ \hline
1	&	1	&	1	&	0	&	-1	&	0	&	0	&	0	&	-1	&	-1	&	0	&	1	&	0	&	0	&	0	&	-1	&	0	&	0	&	0	&	-1	\\ \hline
1	&	1	&	1	&	0	&	1	&	0	&	-1	&	-1	&	0	&	-1	&	-1	&	0	&	-1	&	0	&	1	&	1	&	-1	&	0	&	0	&	1	\\ \hline
1	&	1	&	1	&	1	&	-1	&	-1	&	0	&	0	&	1	&	1	&	-1	&	0	&	-1	&	-1	&	1	&	1	&	0	&	1	&	1	&	0	\\ \hline
1	&	1	&	1	&	1	&	1	&	0	&	0	&	-1	&	-1	&	0	&	0	&	1	&	0	&	1	&	0	&	0	&	0	&	0	&	1	&	1	\\ \hline

\end{tabular}
\end{table}

\begin{thebibliography}{30}


\ifx \showCODEN    \undefined \def \showCODEN     #1{\unskip}     \fi
\ifx \showDOI      \undefined \def \showDOI       #1{#1}\fi
\ifx \showISBNx    \undefined \def \showISBNx     #1{\unskip}     \fi
\ifx \showISBNxiii \undefined \def \showISBNxiii  #1{\unskip}     \fi
\ifx \showISSN     \undefined \def \showISSN      #1{\unskip}     \fi
\ifx \showLCCN     \undefined \def \showLCCN      #1{\unskip}     \fi
\ifx \shownote     \undefined \def \shownote      #1{#1}          \fi
\ifx \showarticletitle \undefined \def \showarticletitle #1{#1}   \fi
\ifx \showURL      \undefined \def \showURL       {\relax}        \fi
\providecommand\bibfield[2]{#2}
\providecommand\bibinfo[2]{#2}
\providecommand\natexlab[1]{#1}
\providecommand\showeprint[2][]{arXiv:#2}

\bibitem[\protect\citeauthoryear{Brown, Kairiss, and Keenan}{Brown
  et~al\mbox{.}}{1990}]%
        {brown1990hebbian}
\bibfield{author}{\bibinfo{person}{Thomas~H Brown}, \bibinfo{person}{Edward~W
  Kairiss}, {and} \bibinfo{person}{Claude~L Keenan}.}
  \bibinfo{year}{1990}\natexlab{}.
\newblock \showarticletitle{Hebbian synapses: biophysical mechanisms and
  algorithms}.
\newblock \bibinfo{journal}{{\em Annual review of neuroscience\/}}
  \bibinfo{volume}{13}, \bibinfo{number}{1} (\bibinfo{year}{1990}),
  \bibinfo{pages}{475--511}.
\newblock


\bibitem[\protect\citeauthoryear{Coleman and Blair}{Coleman and Blair}{2012}]%
        {coleman2012evolving}
\bibfield{author}{\bibinfo{person}{Oliver~J. Coleman} {and}
  \bibinfo{person}{Alan~D. Blair}.} \bibinfo{year}{2012}\natexlab{}.
\newblock \showarticletitle{Evolving Plastic Neural Networks for Online
  Learning: Review and Future Directions}. In \bibinfo{booktitle}{{\em AI 2012:
  Advances in Artificial Intelligence}},
  \bibfield{editor}{\bibinfo{person}{Michael Thielscher} {and}
  \bibinfo{person}{Dongmo Zhang}} (Eds.). \bibinfo{publisher}{Springer Berlin
  Heidelberg}, \bibinfo{address}{Berlin, Heidelberg},
  \bibinfo{pages}{326--337}.
\newblock
\showISBNx{978-3-642-35101-3}


\bibitem[\protect\citeauthoryear{De~Castro}{De~Castro}{2006}]%
        {de2006fundamentals}
\bibfield{author}{\bibinfo{person}{Leandro~Nunes De~Castro}.}
  \bibinfo{year}{2006}\natexlab{}.
\newblock \bibinfo{booktitle}{{\em {Fundamentals of natural computing: basic
  concepts, algorithms, and applications}}}.
\newblock \bibinfo{publisher}{CRC Press}, \bibinfo{address}{Boca Raton,
  Florida}.
\newblock


\bibitem[\protect\citeauthoryear{D{\"u}rr, Mattiussi, Soltoggio, and
  Floreano}{D{\"u}rr et~al\mbox{.}}{2008}]%
        {durr2008evolvability}
\bibfield{author}{\bibinfo{person}{Peter D{\"u}rr}, \bibinfo{person}{Claudio
  Mattiussi}, \bibinfo{person}{Andrea Soltoggio}, {and} \bibinfo{person}{Dario
  Floreano}.} \bibinfo{year}{2008}\natexlab{}.
\newblock \showarticletitle{Evolvability of neuromodulated learning for
  robots}. In \bibinfo{booktitle}{{\em Learning and Adaptive Behaviors for
  Robotic Systems, 2008. LAB-RS'08. ECSIS Symposium on}}.
  \bibinfo{publisher}{IEEE}, \bibinfo{address}{New York},
  \bibinfo{pages}{41--46}.
\newblock


\bibitem[\protect\citeauthoryear{El-Boustani, Ip, Breton-Provencher, Knott,
  Okuno, Bito, and Sur}{El-Boustani et~al\mbox{.}}{2018}]%
        {el2018locally}
\bibfield{author}{\bibinfo{person}{Sami El-Boustani},
  \bibinfo{person}{Jacque~PK Ip}, \bibinfo{person}{Vincent Breton-Provencher},
  \bibinfo{person}{Graham~W Knott}, \bibinfo{person}{Hiroyuki Okuno},
  \bibinfo{person}{Haruhiko Bito}, {and} \bibinfo{person}{Mriganka Sur}.}
  \bibinfo{year}{2018}\natexlab{}.
\newblock \showarticletitle{Locally coordinated synaptic plasticity of visual
  cortex neurons in vivo}.
\newblock \bibinfo{journal}{{\em Science\/}} \bibinfo{volume}{360},
  \bibinfo{number}{6395} (\bibinfo{year}{2018}), \bibinfo{pages}{1349--1354}.
\newblock


\bibitem[\protect\citeauthoryear{Floreano, D{\"u}rr, and Mattiussi}{Floreano
  et~al\mbox{.}}{2008}]%
        {floreano2008neuroevolution}
\bibfield{author}{\bibinfo{person}{Dario Floreano}, \bibinfo{person}{Peter
  D{\"u}rr}, {and} \bibinfo{person}{Claudio Mattiussi}.}
  \bibinfo{year}{2008}\natexlab{}.
\newblock \showarticletitle{{Neuroevolution: from architectures to learning}}.
\newblock \bibinfo{journal}{{\em Evolutionary Intelligence\/}}
  \bibinfo{volume}{1}, \bibinfo{number}{1} (\bibinfo{year}{2008}),
  \bibinfo{pages}{47--62}.
\newblock


\bibitem[\protect\citeauthoryear{Floreano and Urzelai}{Floreano and
  Urzelai}{2000}]%
        {floreano2000evolutionary}
\bibfield{author}{\bibinfo{person}{Dario Floreano} {and}
  \bibinfo{person}{Joseba Urzelai}.} \bibinfo{year}{2000}\natexlab{}.
\newblock \showarticletitle{{Evolutionary robots with on-line self-organization
  and behavioral fitness}}.
\newblock \bibinfo{journal}{{\em Neural Networks\/}} \bibinfo{volume}{13},
  \bibinfo{number}{4-5} (\bibinfo{year}{2000}), \bibinfo{pages}{431--443}.
\newblock


\bibitem[\protect\citeauthoryear{Gerstner, Lehmann, Liakoni, Corneil, and
  Brea}{Gerstner et~al\mbox{.}}{2018}]%
        {gerstner2018eligibility}
\bibfield{author}{\bibinfo{person}{Wulfram Gerstner}, \bibinfo{person}{Marco
  Lehmann}, \bibinfo{person}{Vasiliki Liakoni}, \bibinfo{person}{Dane Corneil},
  {and} \bibinfo{person}{Johanni Brea}.} \bibinfo{year}{2018}\natexlab{}.
\newblock \showarticletitle{Eligibility Traces and Plasticity on Behavioral
  Time Scales: Experimental Support of NeoHebbian Three-Factor Learning Rules}.
\newblock \bibinfo{journal}{{\em Frontiers in Neural Circuits\/}}
  \bibinfo{volume}{12} (\bibinfo{year}{2018}), \bibinfo{pages}{53}.
\newblock


\bibitem[\protect\citeauthoryear{Hebb}{Hebb}{1949}]%
        {hebb1949}
\bibfield{author}{\bibinfo{person}{Donald~Olding Hebb}.}
  \bibinfo{year}{1949}\natexlab{}.
\newblock \bibinfo{booktitle}{{\em {The organization of behavior: A
  neuropsychological theory}}}.
\newblock \bibinfo{publisher}{Wiley \& Sons}, \bibinfo{address}{New York}.
\newblock


\bibitem[\protect\citeauthoryear{Izhikevich}{Izhikevich}{2007}]%
        {izhikevich2007solving}
\bibfield{author}{\bibinfo{person}{Eugene~M. Izhikevich}.}
  \bibinfo{year}{2007}\natexlab{}.
\newblock \showarticletitle{Solving the Distal Reward Problem through Linkage
  of STDP and Dopamine Signaling}.
\newblock \bibinfo{journal}{{\em Cerebral Cortex\/}} \bibinfo{volume}{17},
  \bibinfo{number}{10} (\bibinfo{year}{2007}), \bibinfo{pages}{2443--2452}.
\newblock


\bibitem[\protect\citeauthoryear{Izquierdo-Torres and Harvey}{Izquierdo-Torres
  and Harvey}{2007}]%
        {izquierdo2007hebbian}
\bibfield{author}{\bibinfo{person}{Eduardo Izquierdo-Torres} {and}
  \bibinfo{person}{Inman Harvey}.} \bibinfo{year}{2007}\natexlab{}.
\newblock \showarticletitle{{Hebbian Learning using Fixed Weight Evolved
  Dynamical `Neural' Networks}}. In \bibinfo{booktitle}{{\em Artificial Life,
  2007. ALIFE'07. IEEE Symposium on}}. \bibinfo{publisher}{IEEE},
  \bibinfo{address}{New York}, \bibinfo{pages}{394--401}.
\newblock


\bibitem[\protect\citeauthoryear{Kowaliw, Bredeche, Chevallier, and
  Doursat}{Kowaliw et~al\mbox{.}}{2014}]%
        {kow2016growing}
\bibfield{author}{\bibinfo{person}{Taras Kowaliw}, \bibinfo{person}{Nicolas
  Bredeche}, \bibinfo{person}{Sylvain Chevallier}, {and}
  \bibinfo{person}{Ren{\'e} Doursat}.} \bibinfo{year}{2014}\natexlab{}.
\newblock \showarticletitle{{Artificial neurogenesis: An introduction and
  selective review}}.
\newblock In \bibinfo{booktitle}{{\em Growing Adaptive Machines}}.
  \bibinfo{publisher}{Springer}, \bibinfo{address}{Berlin, Heidelberg},
  \bibinfo{pages}{1--60}.
\newblock


\bibitem[\protect\citeauthoryear{Kuriscak, Marsalek, Stroffek, and
  Toth}{Kuriscak et~al\mbox{.}}{2015}]%
        {kuriscak2015}
\bibfield{author}{\bibinfo{person}{Eduard Kuriscak}, \bibinfo{person}{Petr
  Marsalek}, \bibinfo{person}{Julius Stroffek}, {and} \bibinfo{person}{Peter~G
  Toth}.} \bibinfo{year}{2015}\natexlab{}.
\newblock \showarticletitle{{Biological context of Hebb learning in artificial
  neural networks, a review}}.
\newblock \bibinfo{journal}{{\em Neurocomputing\/}}  \bibinfo{volume}{152}
  (\bibinfo{year}{2015}), \bibinfo{pages}{27--35}.
\newblock


\bibitem[\protect\citeauthoryear{Mocanu, Mocanu, Stone, Nguyen, Gibescu, and
  Liotta}{Mocanu et~al\mbox{.}}{2018}]%
        {mocanu2018scalable}
\bibfield{author}{\bibinfo{person}{Decebal~Constantin Mocanu},
  \bibinfo{person}{Elena Mocanu}, \bibinfo{person}{Peter Stone},
  \bibinfo{person}{Phuong~H Nguyen}, \bibinfo{person}{Madeleine Gibescu}, {and}
  \bibinfo{person}{Antonio Liotta}.} \bibinfo{year}{2018}\natexlab{}.
\newblock \showarticletitle{Scalable training of artificial neural networks
  with adaptive sparse connectivity inspired by network science}.
\newblock \bibinfo{journal}{{\em Nature Communications\/}} \bibinfo{volume}{9},
  \bibinfo{number}{1} (\bibinfo{year}{2018}), \bibinfo{pages}{2383}.
\newblock


\bibitem[\protect\citeauthoryear{Mouret and Tonelli}{Mouret and
  Tonelli}{2014}]%
        {mouret2014artificial}
\bibfield{author}{\bibinfo{person}{Jean-Baptiste Mouret} {and}
  \bibinfo{person}{Paul Tonelli}.} \bibinfo{year}{2014}\natexlab{}.
\newblock \showarticletitle{Artificial evolution of plastic neural networks: a
  few key concepts}.
\newblock In \bibinfo{booktitle}{{\em Growing Adaptive Machines}}.
  \bibinfo{publisher}{Springer}, \bibinfo{address}{Berlin, Heidelberg},
  \bibinfo{pages}{251--261}.
\newblock


\bibitem[\protect\citeauthoryear{Niv, Joel, Meilijson, and Ruppin}{Niv
  et~al\mbox{.}}{2002}]%
        {niv2002evolution}
\bibfield{author}{\bibinfo{person}{Yael Niv}, \bibinfo{person}{Daphna Joel},
  \bibinfo{person}{Isaac Meilijson}, {and} \bibinfo{person}{Eytan Ruppin}.}
  \bibinfo{year}{2002}\natexlab{}.
\newblock \showarticletitle{Evolution of Reinforcement Learning in Uncertain
  Environments: A Simple Explanation for Complex Foraging Behaviors}.
\newblock \bibinfo{journal}{{\em Adaptive Behavior\/}} \bibinfo{volume}{10},
  \bibinfo{number}{1} (\bibinfo{year}{2002}), \bibinfo{pages}{5--24}.
\newblock
\showISSN{1059-7123}


\bibitem[\protect\citeauthoryear{Orchard and Wang}{Orchard and Wang}{2016}]%
        {orchard2016evolution}
\bibfield{author}{\bibinfo{person}{J. Orchard} {and} \bibinfo{person}{L.
  Wang}.} \bibinfo{year}{2016}\natexlab{}.
\newblock \showarticletitle{The evolution of a generalized neural learning
  rule}. In \bibinfo{booktitle}{{\em 2016 International Joint Conference on
  Neural Networks (IJCNN)}}. \bibinfo{publisher}{IEEE}, \bibinfo{address}{New
  York}, \bibinfo{pages}{4688--4694}.
\newblock
\showISSN{2161-4407}


\bibitem[\protect\citeauthoryear{Risi and Stanley}{Risi and Stanley}{2010}]%
        {risi2010indirectly}
\bibfield{author}{\bibinfo{person}{Sebastian Risi} {and}
  \bibinfo{person}{Kenneth~O. Stanley}.} \bibinfo{year}{2010}\natexlab{}.
\newblock \showarticletitle{Indirectly encoding neural plasticity as a pattern
  of local rules}. In \bibinfo{booktitle}{{\em International Conference on
  Simulation of Adaptive Behavior}}. \bibinfo{publisher}{Springer},
  \bibinfo{address}{Berlin, Heidelberg}, \bibinfo{pages}{533--543}.
\newblock


\bibitem[\protect\citeauthoryear{Runarsson and Jonsson}{Runarsson and
  Jonsson}{2000}]%
        {runarsson2000evolution}
\bibfield{author}{\bibinfo{person}{Thomas~Philip Runarsson} {and}
  \bibinfo{person}{Magnus~Thor Jonsson}.} \bibinfo{year}{2000}\natexlab{}.
\newblock \showarticletitle{{Evolution and design of distributed learning
  rules}}. In \bibinfo{booktitle}{{\em Combinations of Evolutionary Computation
  and Neural Networks, 2000 IEEE Symposium on}}. \bibinfo{publisher}{IEEE},
  \bibinfo{address}{New York}, \bibinfo{pages}{59--63}.
\newblock


\bibitem[\protect\citeauthoryear{Soltoggio, Bullinaria, Mattiussi, D{\"u}rr,
  and Floreano}{Soltoggio et~al\mbox{.}}{2008}]%
        {soltoggio2008evolutionary}
\bibfield{author}{\bibinfo{person}{Andrea Soltoggio}, \bibinfo{person}{John~A
  Bullinaria}, \bibinfo{person}{Claudio Mattiussi}, \bibinfo{person}{Peter
  D{\"u}rr}, {and} \bibinfo{person}{Dario Floreano}.}
  \bibinfo{year}{2008}\natexlab{}.
\newblock \showarticletitle{{Evolutionary advantages of neuromodulated
  plasticity in dynamic, reward-based scenarios}}. In \bibinfo{booktitle}{{\em
  International conference on Artificial Life (Alife XI)}}.
  \bibinfo{publisher}{MIT Press}, \bibinfo{address}{Cambridge, MA},
  \bibinfo{pages}{569--576}.
\newblock


\bibitem[\protect\citeauthoryear{Soltoggio and Stanley}{Soltoggio and
  Stanley}{2012}]%
        {soltoggio2012modulated}
\bibfield{author}{\bibinfo{person}{Andrea Soltoggio} {and}
  \bibinfo{person}{Kenneth~O. Stanley}.} \bibinfo{year}{2012}\natexlab{}.
\newblock \showarticletitle{From modulated Hebbian plasticity to simple
  behavior learning through noise and weight saturation}.
\newblock \bibinfo{journal}{{\em Neural Networks\/}}  \bibinfo{volume}{34}
  (\bibinfo{year}{2012}), \bibinfo{pages}{28--41}.
\newblock


\bibitem[\protect\citeauthoryear{Soltoggio, Stanley, and Risi}{Soltoggio
  et~al\mbox{.}}{2018}]%
        {soltoggio2017born}
\bibfield{author}{\bibinfo{person}{Andrea Soltoggio},
  \bibinfo{person}{Kenneth~O. Stanley}, {and} \bibinfo{person}{Sebastian
  Risi}.} \bibinfo{year}{2018}\natexlab{}.
\newblock \showarticletitle{Born to learn: The inspiration, progress, and
  future of evolved plastic artificial neural networks}.
\newblock \bibinfo{journal}{{\em Neural Networks\/}}  \bibinfo{volume}{108}
  (\bibinfo{year}{2018}), \bibinfo{pages}{48--67}.
\newblock


\bibitem[\protect\citeauthoryear{Soltoggio and Steil}{Soltoggio and
  Steil}{2013}]%
        {soltoggio2013solving}
\bibfield{author}{\bibinfo{person}{Andrea Soltoggio} {and}
  \bibinfo{person}{Jochen~J. Steil}.} \bibinfo{year}{2013}\natexlab{}.
\newblock \showarticletitle{Solving the distal reward problem with rare
  correlations}.
\newblock \bibinfo{journal}{{\em Neural computation\/}} \bibinfo{volume}{25},
  \bibinfo{number}{4} (\bibinfo{year}{2013}), \bibinfo{pages}{940--978}.
\newblock


\bibitem[\protect\citeauthoryear{Stanley}{Stanley}{2007}]%
        {stanley2007compositional}
\bibfield{author}{\bibinfo{person}{Kenneth~O. Stanley}.}
  \bibinfo{year}{2007}\natexlab{}.
\newblock \showarticletitle{Compositional pattern producing networks: A novel
  abstraction of development}.
\newblock \bibinfo{journal}{{\em Genetic programming and evolvable machines\/}}
  \bibinfo{volume}{8}, \bibinfo{number}{2} (\bibinfo{year}{2007}),
  \bibinfo{pages}{131--162}.
\newblock


\bibitem[\protect\citeauthoryear{Stanley, Clune, Lehman, and
  Miikkulainen}{Stanley et~al\mbox{.}}{2019}]%
        {stanley2019}
\bibfield{author}{\bibinfo{person}{Kenneth~O Stanley}, \bibinfo{person}{Jeff
  Clune}, \bibinfo{person}{Joel Lehman}, {and} \bibinfo{person}{Risto
  Miikkulainen}.} \bibinfo{year}{2019}\natexlab{}.
\newblock \showarticletitle{Designing neural networks through neuroevolution}.
\newblock \bibinfo{journal}{{\em Nature Machine Intelligence\/}}
  \bibinfo{volume}{1}, \bibinfo{number}{1} (\bibinfo{year}{2019}),
  \bibinfo{pages}{24--35}.
\newblock


\bibitem[\protect\citeauthoryear{Tonelli and Mouret}{Tonelli and
  Mouret}{2013}]%
        {tonelli2013relationships}
\bibfield{author}{\bibinfo{person}{Paul Tonelli} {and}
  \bibinfo{person}{Jean-Baptiste Mouret}.} \bibinfo{year}{2013}\natexlab{}.
\newblock \showarticletitle{On the relationships between generative encodings,
  regularity, and learning abilities when evolving plastic artificial neural
  networks}.
\newblock \bibinfo{journal}{{\em PloS one\/}} \bibinfo{volume}{8},
  \bibinfo{number}{11} (\bibinfo{year}{2013}), \bibinfo{pages}{e79138}.
\newblock


\bibitem[\protect\citeauthoryear{Vasilkoski, Ames, Chandler, Gorchetchnikov,
  L{\'e}veill{\'e}, Livitz, Mingolla, and Versace}{Vasilkoski
  et~al\mbox{.}}{2011}]%
        {vasilkoski2011review}
\bibfield{author}{\bibinfo{person}{Zlatko Vasilkoski}, \bibinfo{person}{Heather
  Ames}, \bibinfo{person}{Ben Chandler}, \bibinfo{person}{Anatoli
  Gorchetchnikov}, \bibinfo{person}{Jasmin L{\'e}veill{\'e}},
  \bibinfo{person}{Gennady Livitz}, \bibinfo{person}{Ennio Mingolla}, {and}
  \bibinfo{person}{Massimiliano Versace}.} \bibinfo{year}{2011}\natexlab{}.
\newblock \showarticletitle{{Review of stability properties of neural
  plasticity rules for implementation on memristive neuromorphic hardware}}. In
  \bibinfo{booktitle}{{\em International Joint Conference on Neural Networks
  (IJCNN)}}. \bibinfo{publisher}{IEEE}, \bibinfo{address}{New York},
  \bibinfo{pages}{2563--2569}.
\newblock


\bibitem[\protect\citeauthoryear{Wilcoxon}{Wilcoxon}{1945}]%
        {wilcoxon1945}
\bibfield{author}{\bibinfo{person}{Frank Wilcoxon}.}
  \bibinfo{year}{1945}\natexlab{}.
\newblock \showarticletitle{Individual comparisons by ranking methods}.
\newblock \bibinfo{journal}{{\em Biometrics bulletin\/}} \bibinfo{volume}{1},
  \bibinfo{number}{6} (\bibinfo{year}{1945}), \bibinfo{pages}{80--83}.
\newblock


\bibitem[\protect\citeauthoryear{Yaman, Mocanu, Iacca, Fletcher, and
  Pechenizkiy}{Yaman et~al\mbox{.}}{2018}]%
        {yaman2018limited}
\bibfield{author}{\bibinfo{person}{Anil Yaman},
  \bibinfo{person}{Decebal~Constantin Mocanu}, \bibinfo{person}{Giovanni
  Iacca}, \bibinfo{person}{George Fletcher}, {and} \bibinfo{person}{Mykola
  Pechenizkiy}.} \bibinfo{year}{2018}\natexlab{}.
\newblock \showarticletitle{Limited evaluation cooperative co-evolutionary
  differential evolution for large-scale neuroevolution}. In
  \bibinfo{booktitle}{{\em Genetic and Evolutionary Computation Conference,
  15-19 July 2018, Kyoto, Japan}}. \bibinfo{publisher}{ACM},
  \bibinfo{address}{New York, NY, USA}, \bibinfo{pages}{569--576}.
\newblock


\bibitem[\protect\citeauthoryear{Zeng and Church}{Zeng and Church}{2009}]%
        {zeng2009finding}
\bibfield{author}{\bibinfo{person}{Wei Zeng} {and} \bibinfo{person}{Richard~L
  Church}.} \bibinfo{year}{2009}\natexlab{}.
\newblock \showarticletitle{Finding shortest paths on real road networks: the
  case for A}.
\newblock \bibinfo{journal}{{\em International journal of geographical
  information science\/}} \bibinfo{volume}{23}, \bibinfo{number}{4}
  (\bibinfo{year}{2009}), \bibinfo{pages}{531--543}.
\newblock


\end{thebibliography}
\end{document}